\documentclass[10pt,twocolumn,letterpaper]{article}

\usepackage{cvpr}              
\usepackage{xcolor}

\definecolor{green}{RGB}{0,183,255}
\definecolor{red}{RGB}{255,14,14}
\newcommand{\std}[1]{{\tiny$\pm$#1}}

\usepackage{multirow}
\usepackage{pifont}
\usepackage{rotating}
\usepackage{array}
\usepackage{booktabs}
\usepackage{adjustbox}
\usepackage{makecell}
\usepackage{listings}
\usepackage{float}
\definecolor{codegreen}{rgb}{0,0.6,0}
\definecolor{codegray}{rgb}{0.5,0.5,0.5}
\definecolor{codepurple}{rgb}{0.58,0,0.82}
\definecolor{backcolour}{rgb}{0.95,0.95,0.92}
\usepackage[hyphens]{url}
\usepackage{xurl}

\definecolor{mygreen}{RGB}{0, 175, 0}

\lstdefinestyle{mystyle}{
    backgroundcolor=\color{backcolour},   
    commentstyle=\color{codegreen},
    keywordstyle=\color{magenta},
    numberstyle=\tiny\color{codegray},
    stringstyle=\color{codepurple},
    basicstyle=\ttfamily\footnotesize,
    breakatwhitespace=false,         
    breaklines=true,                 
    captionpos=b,                    
    keepspaces=true,                 
    numbers=left,                    
    numbersep=5pt,                  
    showspaces=false,                
    showstringspaces=false,
    showtabs=false,                  
    tabsize=2
}

\usepackage{algorithm}
\usepackage{algorithmicx}
\usepackage{etoolbox}
\usepackage{algpseudocode}
\algnewcommand\algorithmicinput{\textbf{Input:}}
\algnewcommand\INPUT{\item[\algorithmicinput]}
\algnewcommand\algorithmicoutput{\textbf{Output:}}
\algnewcommand\OUTPUT{\item[\algorithmicoutput]}

\newcommand{\algstrut}[1][\algruledefaultfactor]{\vrule width 0pt
depth .25\baselineskip height #1\baselineskip\relax}
\newcommand*{\algrule}[1][\algorithmicindent]{\hspace*{.5em}\vrule\algstrut
\hspace*{\dimexpr#1-.5em}}
\makeatletter
\newcount\ALG@printindent@tempcnta
\def\ALG@printindent{%
    \ifnum \theALG@nested>0
    \ifx\ALG@text\ALG@x@notext
    \else
    \unskip
    \ALG@printindent@tempcnta=1
    \loop
    \algrule[\csname ALG@ind@\the\ALG@printindent@tempcnta\endcsname]%
    \advance \ALG@printindent@tempcnta 1
    \ifnum \ALG@printindent@tempcnta<\numexpr\theALG@nested+1\relax
    \repeat
    \fi
    \fi
}%
\patchcmd{\ALG@doentity}{\noindent\hskip\ALG@tlm}{\ALG@printindent}{}{\errmessage{failed to patch}}
\AtBeginEnvironment{algorithmic}{\lineskip0pt}

\newcommand{\thickvert}{\hspace{0.05cm}\raisebox{-0.5ex}{\rule{0.5ex}{2ex}}\hspace{0.1cm}}

\definecolor{cvprblue}{rgb}{0.21,0.49,0.74}
\usepackage[pagebackref,breaklinks,colorlinks,allcolors=cvprblue]{hyperref}

\def\paperID{****} 
\def\confName{CVPR}
\def\confYear{2026}

\title{Towards High-Quality Image Segmentation: Improving Topology Accuracy by Penalizing Neighbor Pixels}

\author{
Juan Miguel Valverde \quad
Dim P. Papadopoulos \quad
Rasmus Larsen \quad
Anders Bjorholm Dahl \\
Technical University of Denmark\\
{\tt\small \{jmvma,dimp,rlar,abda\}@dtu.dk}
}

\begin{document}
\maketitle

\begin{abstract}
Standard deep learning models for image segmentation cannot guarantee topology accuracy, failing to preserve the correct number of connected components or structures.
This, in turn, affects the quality of the segmentations and compromises the reliability of the subsequent quantification analyses.
Previous works have proposed to enhance topology accuracy with specialized frameworks, architectures, and loss functions.
However, these methods are often cumbersome to integrate into existing training pipelines, they are computationally very expensive, or they are restricted to structures with tubular morphology.
We present SCNP, an efficient method that improves topology accuracy by penalizing the logits with their poorest-classified neighbor, forcing the model to improve the prediction at the pixels' neighbors before allowing it to improve the pixels themselves.
We show the effectiveness of SCNP across 13 datasets, covering different structure morphologies and image modalities, and integrate it into three frameworks for semantic and instance segmentation.
Additionally, we show that SCNP can be integrated into several loss functions, making them improve topology accuracy.
Our code can be found at \url{https://jmlipman.github.io/SCNP-SameClassNeighborPenalization}.

\end{abstract}    
\section{Introduction}

\begin{figure} 
  \centering
  \includegraphics[width=0.5\textwidth]{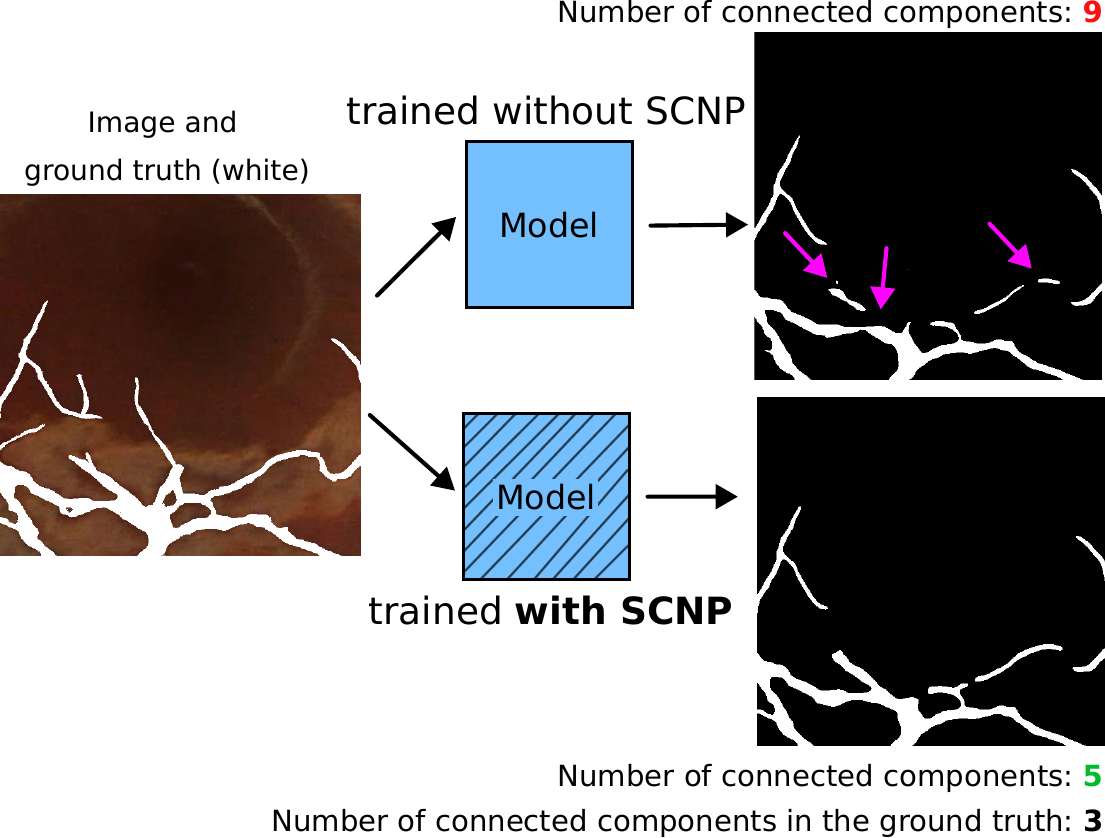}
  \caption{SCNP improves topology accuracy. Arrows indicate breakage in the structures.} \label{fig:teaser}
\end{figure}

Deep learning has become the de-facto standard in image segmentation, yet its standard inference process, which often treats pixels independently, inherently fails to ensure topology accuracy.
This results in the breakage of thin, tubular structures and in small isolated spurious regions (false positives), leading to errors in the number of connected components and, thus, deteriorating image segmentation quality (see \Cref{fig:teaser}).
This shortcoming limits the application of deep learning image segmentation models in many areas.
In fields such as microscopy and medical imaging, where the analysis relies on accurate counts of structures such as cells or axons, or in satellite imagery, where the connectivity of roads is a known prior, it is essential to obtain topologically accurate segmentations.
While achieving topologically perfect segmentations often requires post-processing, models that produce fewer topological errors significantly decrease the burden of manual correction.

Prior work has investigated several research directions to enhance topology accuracy, including specialized architectures or modules \cite{hu2025topopis,song2025optimized,qi2023dynamic,zhao2025deformcl,zhou2025glcp,huang2024representing}, deep learning-based post-processing methods \cite{carneiro2024restoring}, data augmentation \cite{valverde2025disconnect}, and loss functions \cite{pantoja2022topo,hu2019topology,hu2022structure,shit2021cldice,shi2024centerline,acebes2024centerline,kirchhoff2024skeleton,oner2021promoting,stucki2023topologically,zhang2023topology}.
Loss functions offer an elegant and practical solution: they are architecture-agnostic, simple to incorporate, and they require no additional steps after inference.
Most topology loss functions are, however, computationally very expensive or only work on tubular structures.
Persistence homology (PH)-based \cite{edelsbrunner2002topological} topology loss functions (\eg, \cite{hu2019topology,stucki2023topologically,oner2021promoting}) need to compute PH during the optimization, increasing the training time dramatically, from hours to days.
Skeletonization-based topology loss functions (\eg, \cite{shit2021cldice,shi2024centerline,acebes2024centerline}) are CPU-efficient alternatives, but they are unsuitable for datasets with non-tubular structures, such as cells, organs, and brain lesions.
Furthermore, the skeletonization-based clDice loss \cite{shit2021cldice}, which has been shown to be very effective compared to many other topology loss functions \cite{valverde2025topomortar}, demands a large GPU memory and it requires tuning a hyper-parameter that determines its effectiveness.
In summary, to date, there is no plug-and-play, CPU and GPU efficient method that has been demonstrated to improve topology accuracy in datasets with structures of varied morphologies.

In this paper, we present Same Class Neighbor Penalization (SCNP), a method for improving topology accuracy that can be easily incorporated into existing training pipelines, as, in practice, it only requires adding three lines of code (see lines 3-5 in \Cref{alg:scnp} and \Cref{app:code}).
SCNP alters the gradients of the logits by penalizing the logits based on their neighborhood.
Specifically, SCNP penalizes several times the poorest classified predictions around the neighborhood of each logit, forcing the model to improve ``bad'' neighbors before allowing it to improve better pixel predictions.
Additionally, unlike topology loss functions, SCNP is CPU and GPU efficient, requiring only a few extra GPU MBs and milliseconds per iteration (\Cref{app:resources}).

\noindent
Our contributions are:

\begin{itemize}
    \item We present SCNP, a novel method to improve topology accuracy that is efficient and simple to integrate into training pipelines without modifying models' architectures and loss functions.
    \item We provide a theoretical motivation and practical intuition for our SCNP method, demonstrating the advantageous properties of the gradients it generates.
    \item We show on 13 datasets and three frameworks for semantic and instance segmentation that optimizing Cross Entropy Dice loss with SCNP generally yields more accurate segmentations topologically than several topology and non-topology loss functions, which are computationally more expensive and/or require expensive hyper-parameter tuning.
    \item We show that the only hyperparameter of SCNP is intuitive to tune.
    \item We show that SCNP can be incorporated into several standard loss functions, including Cross Entropy, Dice \cite{milletari2016v}, Tversky \cite{salehi2017tversky}, clDice \cite{shit2021cldice}, and Focal loss \cite{lin2017focal}, making them produce segmentations topologically more accurate.
\end{itemize}

\section{Related work}
In image segmentation, not all misclassifications affect segmentation quality equally, and accounting for their importance during training is a central challenge.
Prior work addresses this by re-weighting the loss according to semantic class (\eg, Dice loss \cite{milletari2016v}, weighted Cross Entropy \cite{ronneberger2015u}), error type (\eg, Tversky loss \cite{salehi2017tversky}), or spatial location/context within the image (\eg, RWLoss \cite{valverde2023region}, Boundary loss \cite{kervadec2019boundary}).
Additionally, and as the central focus of this paper, misclassification importance can be weighted based on whether the pixels are critical to the topology of the objects or structures to be segmented, affecting their total number of connected components, holes, and cavities.
Such category of loss functions is referred to as ``topology loss functions'', which can be subdivided into two groups depending on whether they rely on persistence homology.

\noindent
\textbf{Persistence homology} (PH) \cite{edelsbrunner2002topological} is a technique from topological data analysis used to quantify the birth and death of topological features (connected components, holes, and cavities).
PH does this through a ``filtration'', a step-by-step method for gradually constructing a topological space to observe how these features appear and disappear.
The resulting PH information is summarized in a persistence diagram, which is a scatter plot where each point represents a single topological feature, with its coordinates indicating the intensity at which it was born and died.
Topology loss functions employ this framework to identify which pixels are critical for preserving the correct topological structure by applying a filtration to the softmax/sigmoid normalized outputs of a neural network.
\citet{hu2019topology} presented TopoLoss, one of the first topology loss functions employing PH.
\citet{oner2023persistent} proposed a new filtration method to allow PH to capture not only global topological correctness but also the local structure and connectivity of predicted delineations.
\citet{stucki2023topologically} introduced Betti Matching, an alternative to the Wasserstein distance for comparing persistence diagrams.
\citet{xu2024semi} proposed to decompose the topological features in the persistence diagram into noisy and non-noisy, and introduced a loss term to remove the noisy features.
\citet{zhang2025toposculpt} focused on test-time refinement by iteratively improving the segmentation masks based on the a-priori known number of topological features that the segmentations must have.
Due to the reliance on PH, all these methods are extremely expensive to compute, increasing the iteration time from a few milliseconds to several seconds depending on the image size (\Cref{app:resources}).
Furthermore, to partially overcome this issue, most of these methods only compute PH on very small image patches, disregarding global topology.
Our method is, in contrast, CPU efficient, and it can be computed on entire images.

\noindent
\textbf{Skeletonization}-based topology loss functions do not rely on PH homology, focusing on improving the accuracy in or around the skeleton of the structures.
\citet{shit2021cldice} proposed a loss function called clDice, inspired by Dice loss, that involves comparing the skeleton of the prediction against the full ground truth volume, and the skeleton of the ground truth against the full prediction volume.
To compute these skeletons, \citet{shit2021cldice} introduced a differentiable ``soft-skeletonization'' technique.
\citet{viti2022coronary} modified clDice's soft-skeletonization to ensure that the topological features in the ground truth remain connected.
\citet{shi2024centerline,liu2024enhancing} included in the penalization of skeleton pixels the distance to the mask (as in RWLoss \cite{valverde2023region} and Boundary loss \cite{kervadec2019boundary}) and the skeleton thickness.
\citet{kirchhoff2024skeleton} introduced Skeleton Recall (SkelRecall), a computationally efficient alternative to clDice that eliminates the need to compute the prediction’s skeleton, thereby reducing GPU memory usage. SkelRecall focuses on measuring the recall on a dilated version of the ground-truth skeleton, making it particularly effective when training with under-segmented noisy labels \cite{valverde2025topomortar}.
\citet{zhang2023topology} proposed to find critical pixels, that change the topology of the structures, with either a skeletonization method or PH to, ultimately, penalize those critical pixels with MSE loss.
The major disadvantages of skeletonization-based loss functions is that they are restricted to tubular structures. 
In contrast, our method works on structures of any shape.

\noindent
\textbf{Neighbor}-aware methods that consider pixels' neighboring regions represent another research direction relevant to our work.
\citet{rota2017loss} proposed to apply Max Pooling to the per-pixel Cross-Entropy loss map to amplify the worst pixel misclassifications.
\citet{wu2019improving} incorporated an additional branch during training to predict pixel ``affinities'' alongside the primary segmentation maps.
These affinities, which are binary matrices indicating whether two pixels belong to the same class, require selecting an appropriate neighborhood size to exclude pixels that are either trivially close (likely the same class) or too distant (likely unrelated).
\citet{yuan2021neighborloss} introduced NeighborLoss, which penalizes individual pixels in proportion to the number of neighbors classified as a different class, regardless of whether such classification is correct.
\citet{fu2024projected} proposed to magnify small structures with Max Pooling on the planar projections of both the prediction and ground truth.
To further enhance the delineation of fine details and boundary smoothness, total variation \cite{rudin1992nonlinear} is often employed as a regularization term to encourage smooth, well-defined edges (see, \eg, \cite{rabiee2025total}).
Another relevant line of research are Conditional Random Fields (CRFs) methods \cite{lafferty2001conditional}, which explicitly model contextual dependencies among pixels and they can be either applied as a post-processing step \cite{chen2017deeplab} or integrated into training \cite{zheng2015conditional}.
Our method can be used jointly with existing architectures, it can be incorporated to any loss function, and it does not rely on affinity matrices or CRFs, which increase complexity and computational cost.

\section{Method}
\begin{figure} 
  \centering
  \includegraphics[width=0.45\textwidth]{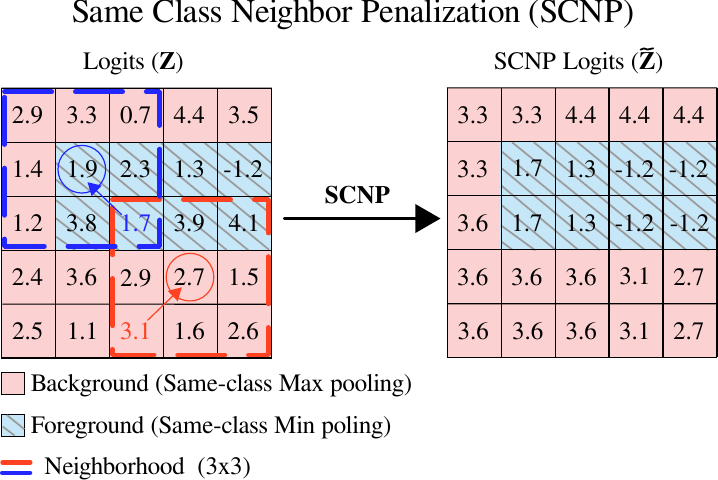}
  \caption{Logits (left) and logits after applying SCNP (right). SCNP propagates small values across the foreground-class logits, and large values across the background-class logits.} \label{fig:method}
\end{figure}

In this section, we introduce Same Class Neighbor Penalization (SCNP), an efficient method to improve topology accuracy that penalizes the logits by propagating their worst neighbors, \ie, the poorest-classified neighboring logits (see \Cref{fig:method}).
This propagation of the worst neighbors, that occurs only during training, forces the models to improve a pixel's neighbors before allowing it to improve the pixel itself.
The \textbf{rationale} and motivation behind SCNP's strategy stems from an observation: the pixels corresponding to small structure breakages and isolated false positives are, necessarily, the worst neighbors of their neighbors.
Thus, improving the accuracy of logits' poorest-predicted neighbors will lead to improving topology accuracy.
SCNP achieves this by selectively propagating the gradients of the pixels with the poorest predictions instead of propagating the gradients of all pixels indiscriminately, as the standard optimization of loss functions does by default.
Throughout the paper, we indicate that a loss function optimized the SCNP-penalized logits with an upper bar, \eg, $\mathcal{L}_{\mathbf{\overline{CE}}}$ or $\mathbf{\overline{CE}}$.

\subsection{Definition}
Let $\mathbf{Z} \in \mathbb{R}^{B \times C \times H \times W}$ be the unnormalized output of a neural network (\ie, the logits) and let $\mathbf{Y} \in \{0,1\}^{B \times C \times H \times W}$ be the one hot-encoded ground truth.
For simplicity, let $\mathbf{z} \in \mathbb{R}^{C \times H W}$ and $\mathbf{y} \in \{0,1\}^{C \times H W}$ denote a single batch instance of $\mathbf{Z}$ and $\mathbf{Y}$, respectively, with the logit $z_{ki}$ corresponding to channel/class $k$ at pixel $i$.
SCNP is defined as:

\begin{equation} \label{eq:scnp_1}
\tilde{z}_{ki} = \left\{ 
\begin{array}{l}
    \underset{j \in \Omega(i) \mid y_{kj}=1}{\min} (z_{kj}) \text{ if } y_{ki} = 1\\
    \underset{j \in \Omega(i) \mid y_{kj}=0}{\max} (z_{kj}) \text{ if } y_{ki} = 0\\
\end{array}
\right. \text{ ,}
\end{equation}
where $\Omega(i)$ represents the neighbor pixels/voxels of $z_{ki}$.
For each channel in $\mathbf{Z}$, SCNP makes each logit corresponding to the foreground class (\ie, if $y_i = 1$ in \cref{eq:scnp_1}) take the smallest value among its foreground-class neighbors, and, simultaneously, SCNP makes each logit corresponding to the background class (\ie, if $y_i = 0$ in \cref{eq:scnp_1}) take the largest value among its background-class neighbors.
After computing the SCNP logits (\ie, $\mathbf{\tilde{Z}} = \text{SCNP}(\mathbf{Z})$), optimization is performed in the standard way, by simply minimizing  $\mathcal{L}(\sigma(\mathbf{\tilde{Z}}), \mathbf{Y})$, where $\mathcal{L}$ is a loss function (\eg, Cross Entropy, or Dice loss), and $\sigma$ is an activation function (sigmoid, softmax).

Optimizing a loss function with the SCNP logits $\mathbf{\tilde{Z}}$ produces three key effects.
\textbf{First}, SCNP increases the loss, as it deteriorates the logits which, in turn, deteriorates their softmax/sigmoid probability values.
As illustrated in \Cref{fig:contribution}, SCNP acts on the center pixel's logits corresponding to the ground truth class 0, deteriorating them from $[2.3, 1.2, 1.4]$ to $[1.9, 1.7, 1.7]$, which also deteriorates the softmax probability values from $[0.57, 0.19, 0.23]$ to $[0.38, 0.31, 0.31]$.
\textbf{Second}, the logits propagated with SCNP that replaced other logits (\ie, the logits corresponding to the poorest predictions) will be penalized as many times as they were propagated.
Note how, for instance, the logit 1.2 in \Cref{fig:contribution} (Channel 2, yellow) is propagated three times (Channel 2, blue).
In consequence, in order to improve pixel predictions at those three pixel locations, the model has to first improve the prediction at the pixel corresponding to the logit that was propagated, and since the loss is computed per pixel, its computation will then involve the logit that was propagated three times.
The \textbf{third} effect is that the gradients produced by SCNP couple predictions at different neighbor pixels and across classes.

\subsection{Gradients}

\begin{figure} 
  \centering
  \includegraphics[width=0.5\textwidth]{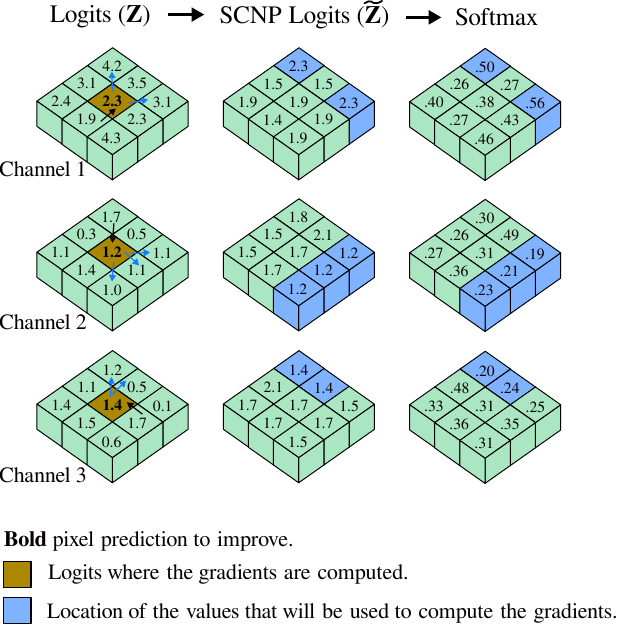}
  \caption{Logits after applying SCNP and softmax normalization.} \label{fig:contribution}
\end{figure}

The gradients of a loss function \wrt the logits can be computed as: $\frac{\partial \mathcal{L}}{\partial z_{ki}} = \frac{\partial \mathcal{L}}{\partial \hat{y}_{ki}} \cdot \frac{\partial \hat{y}_{ki}}{\partial z_{ki}}$.
Employing the SCNP logits adds the following term to the chain rule:

\begin{equation} \label{eq:scnp}
    \frac{\partial \tilde{z}_{ki} }{\partial z_{ki}} = \sum_{j \in \Omega(i)} \mathbf{1}[z_{ki} = \tilde{z}_{kj}] \frac{\partial \mathcal{L}}{\partial \tilde{z}_{kj}},
\end{equation}
where $\mathbf{1}[\cdot]$ is 1 if the inner condition is true and 0 otherwise.
\Cref{eq:scnp} shows that 1) the poorest misclassifications are penalized as many times as they are propagated by SCNP, and that 2) SCNP intertwines predictions at neighboring pixels across classes.
\Cref{fig:contribution} illustrates this: computing the gradients of the loss \wrt the center pixel (yellow logits, left column) involves the values from two neighbor pixels at Channel 1, three neighbor pixels at Channel 2, and two neighbor pixels at Channel 3 (blue softmax probabilities, right column).
\Cref{app:gradients} revisits the gradients of various loss functions, and it provides an example of the computation of the gradients.

SCNP, by propagating the logits' worst neighbors, inhibits the propagation of the gradients corresponding to the other logits.
In other words, SCNP limits the number of pixel predictions that are directly improved in each iteration by focusing on improving only the worst neighbors.
To illustrate, consider the logits $[3.1, 0.3, 1.1]$ in \Cref{fig:contribution} (Channels 1,2,3, respectively); since these three logits are replaced by others after applying SCNP ($[1.5, 1.5, 2.1]$), the gradients corresponding to the original pixel prediction $[3.1, 0.3, 1.1]$ will not be propagated during training.
Optimizing the SCNP-penalized logits can be, thus, interpreted as optimizing an upper bound of the loss, which can slow down convergence.
We observed that, in practice, this is not an issue, as optimizing the SCNP logits results in a negligible delay in convergence that, if needed, can be reduced by optimizing the standard logits and the SCNP-penalized logits jointly, \ie, $\mathcal{L}(\sigma(\mathbf{\tilde{Z}}), \mathbf{Y}) + \mathcal{L}(\sigma(\mathbf{Z}), \mathbf{Y})$. More details can be found in \Cref{app:convergence}.

\subsection{Implementation}

SCNP can be efficiently implemented with Max and Min Pooling by masking out the foreground and background logits for each class separately.
This can be achieved by multiplying the background logits of each class with a very large positive number $\kappa$ when computing Min Pooling, and by multiplying the foreground logits with a very large negative number $-\kappa$ when computing Max Pooling (see \Cref{alg:scnp}).
The logits multiplied by the large positive and negative number will, consequently, not be propagated to the background/foreground class when applying Min/Max Pooling.
The use of Pooling introduces a hyper-parameter, window size ($w$), that determines the size of the neighborhood that will be considered when propagating the logits.
Throughout this paper, we fix the neighborhood size at $w=3$ unless explicitly stated otherwise.
This setting provides favorable results in practice; nevertheless, the performance of SCNP may be further improved by finding the optimal $w$ for a specific dataset.
The other hyper-parameters introduced by the Pooling layers are always fixed: stride is always 1, and the padding is set such that the input and output feature maps have the same size.

The operations utilized by SCNP (\ie, Pooling and matrix algebraic operations) barely demand additional resources, increasing running time by only a few milliseconds per iterations and GPU memory usage by just a few MiBs (see details in \Cref{app:resources}).
Furthermore, SCNP can be conveniently wrapped as a function that is applied after the computation of the logits and before the computation of the loss function (see the code in \Cref{app:code}).

\begin{algorithm}[t]
\caption{SCNP}
\label{alg:scnp}
\begin{algorithmic}[1]
\INPUT logits $\mathbf{Z}$, one hot-encoded ground truth $\mathbf{Y}$.
\State  \textbf{Given}: Kernel size $w$ (default: $w=3$).
\State  \textbf{Set}: $\kappa=\infty$ \textit{\# Any very large number}
\State $\mathbf{V}_1 \leftarrow MinPool((\mathbf{Z} \odot \mathbf{Y}) + \kappa(1-\mathbf{Y}), w)$
\State $\mathbf{V}_2 \leftarrow MaxPool((\mathbf{Z} \odot (1-\mathbf{Y})) - \kappa\mathbf{Y}, w)$
\State $\tilde{\mathbf{Z}} \leftarrow (\mathbf{V}_1 \odot \mathbf{Y}) + (\mathbf{V}_2 \odot (1 - \mathbf{Y}))$

\OUTPUT SCNP logits $\tilde{\mathbf{Z}}$
\end{algorithmic}
\end{algorithm}

\section{Experiments}
We conducted three experiments: 1) An extensive benchmark comparing Cross Entropy Dice loss with SCNP against several loss functions on various datasets and frameworks, 2) a sensitivity analysis on the only hyper-parameter of SCNP, and 3) an ablation study where we used SCNP jointly with several loss functions.
We also compared SCNP with binary closing (\Cref{app:binaryclosing}).

\subsection{Benchmark} \label{sec:benchmark}
We evaluated and compared: 1) CEDice optimization applied to the standard logits (our baseline), 2) Cross Entropy Dice loss applied to the SCNP-penalized logits ($\mathcal{L}_{\overline{CEDice}}$), 3) the combination of these two (\ie, $\mathcal{L}_{CEDice+\overline{CEDice}}$), and 4) several other topology and non-topology loss functions.
We chose CEDice as a baseline because it is standard in segmentation tasks where topology accuracy is crucial.

\paragraph{Datasets and models}
We employed 13 datasets and three frameworks.
nnUNetv2 \cite{isensee2021nnu} was used for semantic segmentation on medical datasets with tubular structures (FIVES \cite{jin2022fives}, Axons \cite{abdollahzadeh2021deepacson}, PulmonaryVA \cite{cheng2024fusion}) and non-tubular structures (ATLAS2 \cite{liew2022large}, ISLES24 \cite{de2024isles}, CirrMRI600 \cite{jha2025large}, and MSLesSeg \cite{guarnera2025mslesseg}).
Detectron2/DeepLabv3+ \cite{wu2019detectron2,chen2018encoder} was used for semantic segmentation on non-medical datasets with tubular structures (TopoMortar \cite{valverde2025topomortar}, DeepRoads \cite{demir2018deepglobe}, Crack500 \cite{yang2019feature}).
InstanSeg \cite{goldsborough2024instanseg} was used for instance segmentation on rounded cell datasets (IHC\_TMA \cite{wang2024simultaneously}, LyNSeC \cite{naji2024holy}, NuInsSeg \cite{mahbod2024nuinsseg}).
All datasets had a background and a foreground class, except PulmonaryVA which had two foreground classes.
The final activation function employed in semantic segmentation datasets was softmax while, in instance segmentation datasets, it was sigmoid.
More details about the datasets, their size, and the splits can be found in \Cref{app:datasets}.

\begin{figure*}
  \centering
  \includegraphics[width=\textwidth]{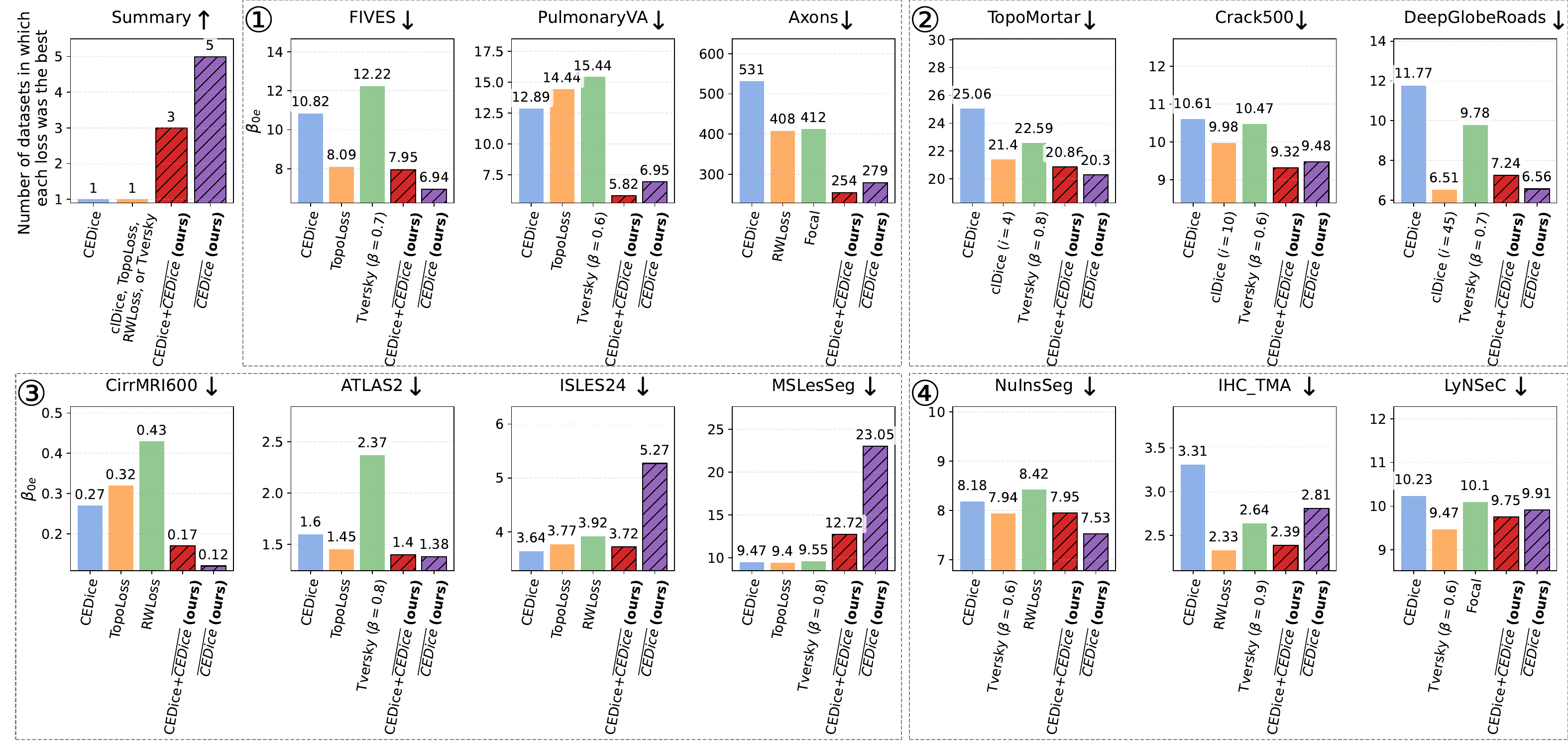}
  \caption{$\beta_{0e}$ obtained by Cross Entropy Dice loss after optimizing the standard logits (CEDice), the SCNP-penalized logits ($\overline{CEDice}$), both (CEDice+$\overline{CEDice}$), and the top two loss functions across the 13 datasets. Datasets are grouped into four categories (\ding{172} \ding{173} \ding{174} \ding{175}) based on their image type (medical, non-medical) and structures' morphologies (tubular, non-tubular, rounded). First panel: Number of datasets in which each loss function led the most accurate segmentations topologically.} \label{fig:barplotresults}
\end{figure*}

\paragraph{Loss functions}
We optimized seven and five loss functions on datasets with tubular and non-tubular structures, respectively, as two losses are constraint to tubular structures.
We optimized Cross Entropy Dice (CEDice), Focal \cite{lin2017focal} and Tversky loss \cite{salehi2017tversky} since they are standard in image segmentation; TopoLoss \cite{hu2019topology}, clDice \cite{shit2021cldice}, and Skeleton Recall loss (SkelRecall) \cite{kirchhoff2024skeleton} are topology loss functions; and RegionWise loss \cite{valverde2023region} (RWLoss) is a distance-based loss function that has been shown to improve topology accuracy \cite{liu2024enhancing}.
For clDice and Tversky loss, we conducted hyper-parameter optimization on each of the 13 datasets, as their performance is very sensitive to the hyper-parameter choice (\Cref{app:loss_functions_extra}).
We did not use Max Pooling loss \cite{rota2017loss} since it is too inefficient for our large-scale benchmark, as it requires sorting all pixel predictions on each training iteration, and we did not use NeighborLoss \cite{yuan2021neighborloss} since it seems to increase false positives and negatives, as it penalizes locally-inconsistent pixel predictions without considering the ground truth.

\paragraph{Metrics}
We measured Dice coefficient \cite{dice1945measures} and the Betti error \cite{hu2019topology}, which is standard to measure topology accuracy.
We report the first Betti error $\beta_{0e}$ (\ie, the difference in the number of connected components between the prediction and the ground truth) as it is the most relevant measure for the datasets used.
On datasets with tubular structures we, additionally, measured clDice \cite{shit2021cldice}, which also reflects topology accuracy, and on dataset with rounded structures, we also measured roundness $(\mathcal{R}_e)$ \cite{pieta2025fast}.
We run every experiment five times.

\begin{figure*} 
  \centering
  \includegraphics[width=\textwidth]{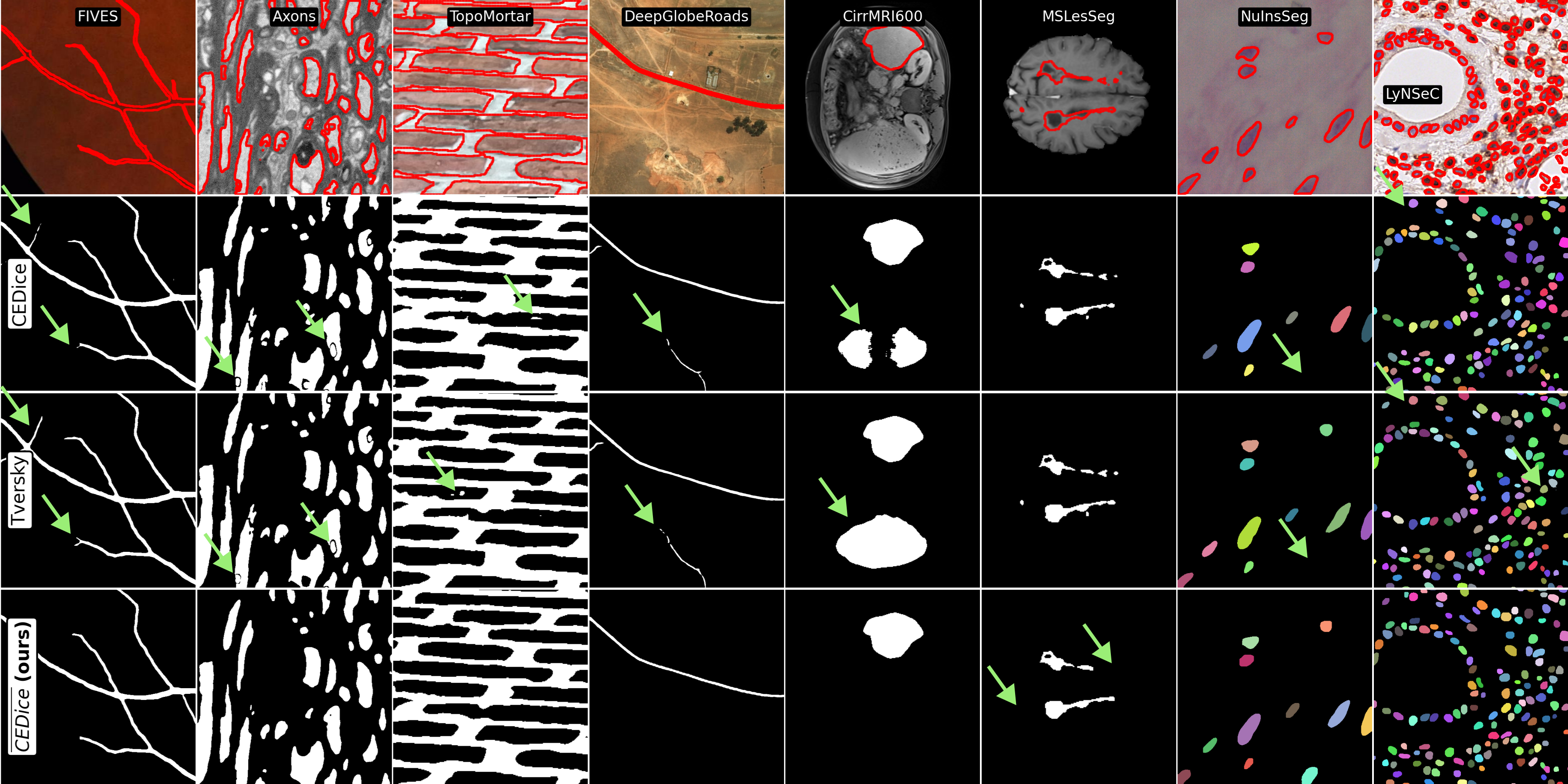}
  \caption{Segmentations achieved by the baseline Cross Entropy Dice loss optimizing the standard logits (CEDice), the SCNP-penalized logits ($\overline{CEDice}$), and Tversky loss (the best-performing loss after CEDice) in eight datasets (two from each dataset group). Arrows indicate topological errors.} \label{fig:main_qualitative_results}
\end{figure*}

\paragraph{Results} 
We organized the 13 datasets into four groups according to image type and structure morphology to provide a detailed analysis of the conditions under which SCNP is advantageous: Group \ding{172} Medical datasets with tubular structures; Group \ding{173} Non-medical datasets with tubular structures; Group \ding{174} Medical datasets with non-tubular structures; Group \ding{175} Medical datasets with rounded structures (cells).
\Cref{fig:barplotresults} shows the average $\beta_{0e}$ provided by SCNP on the four groups of datasets, and \Cref{fig:main_qualitative_results} presents qualitative segmentation results.
Due to space limitations, \Cref{fig:barplotresults} includes only the top two loss functions for each dataset, and \Cref{fig:main_qualitative_results} shows only Tversky loss (the best performing competing loss) and two datasets from each group.
We included the tables with the mean and the standard deviation of the Dice coefficient, $\beta_{0e}$, clDice, and roundness of all datasets on \Cref{app:main_results_tables}, and illustrative examples of segmentations of all datasets in \Cref{app:quality}.

\paragraph{$\vartriangleright$ Group \ding{172} Medical datasets with tubular structures} Our SCNP led the lowest $\beta_{0e}$ in the three datasets while not deteriorating the Dice or the clDice metric.
In most cases, topology and non-topology loss functions (without our SCNP) did not surpass CEDice loss (see \Cref{fig:barplotresults} and \Cref{table:skeleton_medical_nnunetv2} in \Cref{app:main_results_tables}).

\paragraph{$\vartriangleright$ Group \ding{173} Non-medical datasets with tubular structures} On TopoMortar and Crack500, SCNP achieved the lowest $\beta_{0e}$ while not deteriorating Dice and clDice.
On DeepGlobeRoads, SCNP led to better $\beta_{0e}$ than most loss functions at the expense of deteriorating Dice and clDice. On this dataset, the only loss that did not deteriorate significantly the average Dice coefficient and clDice was Tversky loss with an optimal hyper-parameter choice.
On Crack500 and DeepGlobeRoads datasets, RWLoss and Focal loss led to empty segmentations (see \Cref{fig:barplotresults} and \Cref{table:skeleton_natural_detectron} in \Cref{app:main_results_tables}).

\paragraph{$\vartriangleright$ Group \ding{174} Medical datasets with non-tubular structures} The effectiveness of SCNP varied significantly across datasets.
On CirrMRI600, SCNP was clearly beneficial, decreasing the $\beta_{0e}$ by half compared to the baseline and achieving its lowest value.
On MSLesSeg, SCNP was detrimental, especially when optimizing $\mathcal{L}_{\overline{CEDice}}$ alone.
On ATLAS2, SCNP achieved slightly lower $\beta_{0e}$ and Dice coefficient, and on ISLES24 no method, including SCNP, achieved better performance than CEDice, with $\mathcal{L}_{\overline{CEDice}}$ slightly worsening the segmentations (see \Cref{fig:barplotresults} and \Cref{table:round_medical_nnunet} in \Cref{app:main_results_tables}).

\paragraph{$\vartriangleright$ Group \ding{175} Medical datasets with rounded structures (cells)} SCNP achieved the lowest $\beta_{0e}$ on NuInsSeg and IHC\_TMA, and it improve the Dice coefficient on IHC\_TMA.
On LyNSeC, SCNP led to the 2{\textsuperscript {nd}} best results after Tversky loss with its optimal hyper-parameter.
Furthermore, SCNP improved roundness on all datasets (see \Cref{fig:barplotresults} and \Cref{table:round_cells_instance} in \Cref{app:main_results_tables}).

\subsection{Sensitivity analysis on the neighborhood size} \label{sec:sensitivity}

We investigated how SCNP's optimal neighborhood size ($w$) depends on tubular thickness.
To this end, we conducted our experiments on different resized versions of FIVES, ensuring that performance differences were attributable to structure thickness and not to the different dataset characteristics.
First, we down-sampled FIVES four times, resulting in five datasets with image size of $2048^2$ (original), $1536^2$, $1024^2$, $512^2$, and $256^2$ pixels and an average median blood vessel thickness of 9.69, 7.23, 4.70, 1.88, and 1.00 pixels respectively.
Resizing FIVES several times resulted in a significant breakage of the structures' ground truth, with an average number of connected components of 4.4, 5.6, 5.7, 15 and 69, respectively.
Thus, we conducted our experiments on the three highest resolution datasets.

\paragraph{Results} \Cref{table:sensitivity} shows the $\beta_{0e}$ of our experiments (the Dice and clDice scores can be found in \Cref{app:sensitivity_dice_cldice}).
\Cref{table:sensitivity} highlights two key findings.
First, the optimal $w$ value utilized in our benchmark ($w=3$) was suboptimal despite achieving the lowest $\beta_{0e}$ compared to other loss functions (\Cref{fig:barplotresults}, FIVES).
Second, the optimal $w$ corresponded approximately to the blood vessel thickness: with a median blood vessel of 9.69, the best $w$ was 9; with a median blood vessel of 7.23, the best was 7; and with a median of 4.70, the best values were 5 and 7.

\begin{table}[]
\centering
{\small
\begin{tabular}{|l|lll|}
\hline
$\downarrow w$ & $2048^2$ (\thickvert 9.69)  & $1024^2$ (\thickvert 7.23) & $512^2$ (\thickvert 4.70) \\ \hline
3 & 6.94\std{1.47} & 7.12\std{1.80} & 8.57\std{1.98} \\
5 & 5.64\std{1.34} & 5.96\std{1.51} & 7.79\std{1.63} \\
7 & 4.87\std{1.20} & \textbf{5.45\std{1.61}} & \textbf{7.78\std{1.90}} \\
9 & \textbf{4.79\std{1.31}} & 7.14\std{2.86} & 7.89\std{2.34} \\
11 & 7.80\std{1.87} & 11.02\std{2.55} & 18.65\std{3.72} \\
\hline
\end{tabular}
} \caption{Sensitivity analysis on SCNP's hyperparameter ($w$) across three versions of FIVES. Top row: pixel size$^2$, \thickvert: median thickness of the blood vessels.} \label{table:sensitivity}
\end{table}

\subsection{Ablation study} \label{sec:ablation}
We investigated the effectiveness and generalizability of SCNP by optimizing all the loss functions used in this study (including Cross Entropy (CE) and Dice loss, separately) over the SCNP-penalized logits.
We conducted the ablation study on FIVES, and, for the loss functions that required hyper-parameter tuning, we utilized the optimal hyper-parameter value that we found during the benchmark.

\paragraph{Results} SCNP improved the performance of all loss functions, without exception: $\beta_{0e}$ decreased in every case, while Dice and clDice did not worsen and, occasionally, improved (see \Cref{table:ablation}).

\begin{table}[]
\centering
{\small
\begin{tabular}{|l|lll|}
\hline
Loss & Dice & $\beta_{0e}$ & clDice \\ \hline
CE & 0.92\std{0.0} & 11.93\std{2.92} & 0.92\std{0.0} \\ 
$\overline{CE}$ (\textbf{ours}) & 0.92\std{0.0} & \textbf{7.53\std{1.74}} & 0.92\std{0.0} \\ 
\hline
Dice & 0.92\std{0.0} & 12.03\std{3.4} & 0.92\std{0.0} \\ 
$\overline{Dice}$ (\textbf{ours}) & 0.92\std{0.0} & \textbf{7.88\std{2.03}} & 0.92\std{0.0} \\ 
\hline
CEDice & 0.92\std{0.0} & 10.82\std{2.71} & 0.92\std{0.0} \\ 
CEDice+$\overline{CEDice}$ (\textbf{ours}) & 0.92\std{0.0} & \textbf{6.94\std{1.47}} & 0.92\std{0.0} \\ 
\hline
clDice ($i=24$) & 0.91\std{0.0} & 36.55\std{10.70} & 0.91\std{0.01} \\ 
$\overline{clDice}$ ($i=24$) (\textbf{ours}) & \textbf{0.92\std{0.0}} & \textbf{5.44\std{1.91}} & \textbf{0.93\std{0.0}} \\ 
\hline
SkelRecall & 0.91\std{0.0} & 12.45\std{4.25} & 0.91\std{0.0} \\ 
$\overline{SkelRecall}$ (\textbf{ours}) & 0.91\std{0.0} & \textbf{5.07\std{1.91}} & 0.92\std{0.0} \\ 
\hline
TopoLoss & 0.92\std{0.0} & 8.09\std{2.45} & 0.92\std{0.0} \\ 
$\overline{TopoLoss}$ (\textbf{ours}) & 0.92\std{0.0} & \textbf{6.36\std{1.55}} & 0.92\std{0.0} \\ 
\hline
RWLoss & 0.92\std{0.0} & 15.99\std{5.35} & 0.92\std{0.0} \\ 
$\overline{RWLoss}$ (\textbf{ours}) & 0.92\std{0.0} & \textbf{8.64\std{3.36}} & 0.91\std{0.0} \\ 
\hline
Tversky ($\beta = 0.7$) & 0.92\std{0.0} & 12.22\std{3.45} & 0.92\std{0.0} \\ 
$\overline{Tversky}$ ($\beta = 0.7$) (\textbf{ours}) & 0.92\std{0.0} & \textbf{7.31\std{2.20}} & \textbf{0.93\std{0.0}} \\ 
\hline
Focal & 0.92\std{0.0} & 16.08\std{5.97} & 0.92\std{0.0} \\ 
$\overline{Focal}$ (\textbf{ours}) & 0.92\std{0.0} & \textbf{7.75\std{1.58}} & 0.92\std{0.0} \\ 
\hline
\end{tabular}
} \caption{Ablation comparing several loss functions optimizing the standard logits vs. the SCNP-penalized (ours) logits.} \label{table:ablation}
\end{table}

\section{Discussion}

We presented SCNP, an efficient method to improve topology accuracy by penalizing the logits with their worst neighbors, \ie, logits' neighbors with the poorest predictions.
We showed how SCNP gradients allow to incorporate pixel neighbor information during the optimization, and that SCNP can be used with existing architectures, frameworks, and loss functions, requiring only three extra lines of code placed after the computation of the logits and before the loss (\Cref{app:code}).

We benchmarked SCNP on 13 datasets, categorized by image type (medical, non-medical) and structure morphology (tubular, non-tubular, rounded), to determine when SCNP is most beneficial.
On \textbf{medical and non-medical datasets with tubular structures} (see \ding{172} \ding{173} in \Cref{fig:barplotresults} and \Cref{table:skeleton_medical_nnunetv2,table:skeleton_natural_detectron} in \Cref{app:main_results_tables}), SCNP demonstrated a clear advantage, as it made CEDice loss improve topology accuracy, surpassing the other loss functions.
Only on one of the six datasets with tubular structures (DeepGlobeRoads) SCNP improved topology accuracy at the expense of slightly deteriorating the Dice coefficient, a trade-off that is arguably desirable given the critical importance of road network connectivity.
Although only one of the datasets was multi-class, SCNP is designed to work with an arbitrary number of classes.
On \textbf{medical datasets with rounded structures}, SCNP was also very effective, as it improved topology accuracy and, on IHC\_TMA, it improved Dice coefficient as well (see \ding{175} in \Cref{fig:barplotresults} and \Cref{table:round_cells_instance} in \Cref{app:main_results_tables}).
SCNP also improved the roundness of the segmentations in all three datasets, which is advantageous in datasets with structures that are known to be rounded, such as cells.
We attribute this improvement to SCNP's incorporation of neighbor information during the optimization, which likely encourages smooth borders and structural coherence.

On the \textbf{medical datasets with non-tubular structures}, SCNP's effectiveness ranged from highly beneficial to detrimental (see \ding{174} in \Cref{fig:barplotresults} and \Cref{table:round_medical_nnunet} in \Cref{app:main_results_tables}).
We investigated two potential factors contributing to this inconsistency.
First, we considered class imbalance, as the foreground percentage in ATLAS2 (0.38\%), ISLES (0.43\%) and MSLesSeg (0.17\%) was the lowest across all datasets.
We believe this is not the primary cause since Dice loss inherently address class imbalance, and SCNP was very effective in a similarly imbalanced dataset (PulmonaryVA (1.67\% foreground)).
The second factor that we investigated was structure size.
The average structure in CirrMRI600, ATLAS2, ISLES24, and MSLesSeg was 227859, 13071, 7058, and 447 voxels, respectively, which mirrors the descending order of SCNP's observed effectiveness.
We hypothesize that the structural coherence and smooth borders promoted by SCNP, while generally beneficial, may become disadvantageous when segmenting extremely small structures and, particularly, when the contrast between their boundaries and the background is low.

Our sensitivity analysis on the only hyper-parameter of SCNP ($w$) indicates that its optimal value on datasets with tubular structures is correlated to structure thickness.
This is likely because a too small neighborhood will not cover the areas that can lead to structure breakage, while a too large neighborhood includes far, unrelated pixels.
While, based on this observation, the SCNP's hyper-parameter is intuitive to tune, its default value ($w=3$) has been shown generally effective across all the datasets (\Cref{sec:benchmark}).

Our ablation study confirms that SCNP can make several loss functions improve topology accuracy, including standard losses (CE, Dice \cite{milletari2016v}, Tversky \cite{salehi2017tversky}, Focal loss \cite{lin2017focal}) and topology losses (clDice \cite{shit2021cldice}, SkelRecall \cite{kirchhoff2024skeleton}, and TopoLoss \cite{hu2019topology}) (see \Cref{sec:ablation}).
This versatility is one of the main advantages of SCNP compared to related work, including Max Pooling loss \cite{rota2017loss} and NeighborLoss \cite{yuan2021neighborloss}, as it permits the utilization of the loss function that is most suitable for a particular dataset/task (\eg Dice loss in class-imbalanced datasets) while simultaneously improving topology accuracy.
\section{Conclusion}

Penalizing by the worst performing pixel in a neighborhood, as proposed in our SCNP method, is an efficient approach for enhancing topology accuracy that can be readily used in combination with existing deep learning models, loss functions, and training pipelines.
Our exhaustive comparisons revealed that SCNP is effective in most datasets where topology accuracy is important.
We showed that the default value of SCNP's only hyper-parameter is effective, it is intuitive to tune, and that the performance of eight loss functions improved when optimizing the SCNP-penalized logits.

\section*{Acknowledgements}
We thank Thanos Delatolas for conducting some exploratory experiments during the early stage of this project.
This work was supported by NordForsk and a research grant (VIL50087) from VILLUM FONDEN.
{
    \small
    \bibliographystyle{ieeenat_fullname}
    \bibliography{main}
}

\clearpage
\setcounter{page}{1}
\appendix
\maketitlesupplementary
\setcounter{figure}{0} 
\setcounter{table}{0} 
\setcounter{equation}{0} 
\renewcommand{\thefigure}{\Alph{section}.\arabic{figure}}
\renewcommand{\thetable}{\Alph{section}.\arabic{table}}
\crefalias{section}{appendix}
\crefname{appendix}{Appendix}{Appendices}

\section{Implementation (code)} \label{app:code}

The three-lines implementation of SCNP that we employed in this study. Note that $Y$ is the one-hot encoded ground truth, stride is always 1, and the padding is determined by the kernel size. Also note that SCNP only needs to be applied \underline{during training}.

\begin{lstlisting}[language=Python, caption={SCNP implementation.}]
logits = model(X)
# SCNP Begin
t1 = -torch.nn.functional.max_pool2d(-(logits*Y + 9999*(1-Y)), kernel_size=3, stride=1, padding=1)
t2 = torch.nn.functional.max_pool2d((logits*(1-Y) - 9999*Y), kernel_size=3, stride=1, padding=1)
scnp_logits = t1*Y + t2*(1-Y)
# SCNP End
loss = CEDiceLoss(softmax(scnp_logits), Y)
\end{lstlisting}

\section{Computational resources required by loss functions and SCNP} \label{app:resources}

\Cref{table:app_resouces_1,table:app_resouces_2} show the running time of one epoch and the GPU memory required by each loss function and our SCNP.
These benchmarks were run on a cluster node entirely reserved for computing the measurements, consisting of 2x Intel(R) Xeon(R) Gold 6326 @ 2.90GHz and 1x NVIDIA A10 with 24GB.
For SCNP, we computed the difference in resources when optimizing $\mathcal{L}_{CEDice}$ and $\mathcal{L}_{\overline{CEDice}}$.

We benchmarked all methods on FIVES (2D) and PulmonaryVA (3D).
The size of the tensors sent to the model were $2 \times 2 \times 1280 \times 1024$ and $2 \times 3 \times 160 \times 128 \times 112$ on FIVES and PulmonaryVA, respectively.
An epoch consisted on 300 training iterations; on FIVES that amounted to 2100 forward passes, whereas on PulmonaryVA it amounted to 1200 forward passes.
This is due to the use of Deep supervision and different models that were automatically configured by nnUNetv2.

\begin{table}[h]
\centering
{
\begin{tabular}{|l|ll|}
\hline
Loss & Epoch time (sec.) & GPU (MiB) \\ \hline
CEDice & 54.85 & 5350 \\
clDice ($i=24$) & 181.78 & {\color{red} 19340}  \\
SkelRecall & 115.59 & 5334  \\
TopoLoss & {\color{red} 1505} & 5334  \\
RWLoss & 56.59 & 5778  \\
Tversky ($\beta=0.7$) & 54.76 & 5326  \\
Focal & 55.26 & 5678  \\
\hline
\multicolumn{3}{|l|}{Extra resources required by:} \\
\hline
SCNP & \textcolor{mygreen}{+1.47}  & \textcolor{mygreen}{+340} \\
\hline
\end{tabular}
} \caption{Resources required on FIVES (2D nnUNetv2).} \label{table:app_resouces_1}
\end{table}

\begin{table}[h]
\centering
{
\begin{tabular}{|l|ll|}
\hline
Loss & Iteration time (sec.) & GPU (MiB) \\ \hline
CEDice & 72.79 & 6644 \\
clDice ($i=16$) & 199.10 & {\color{red} 38672}  \\
SkelRecall & 157.17 & 7784  \\
TopoLoss & {\color{red} 12960} & 6632  \\
RWLoss & 79.42 & 6864  \\
Tversky ($\beta=0.6$) & 73.02 & 6644  \\
Focal & 70.65 & 6778  \\
\hline
\multicolumn{3}{|l|}{Extra resources required by:} \\
\hline
SCNP & \textcolor{mygreen}{+0.66} & \textcolor{mygreen}{+454}  \\
\hline
\end{tabular}
} \caption{Resources required on PulmonaryVA (3D nnUNetv2).} \label{table:app_resouces_2}
\end{table}

\section{Gradients} \label{app:gradients}
The derivative of a loss function \wrt the logits can be obtained via the chain rule

\begin{equation} \label{eq:chainrule}
\frac{\partial \mathcal{L}}{\partial z_{ki}} = \frac{\partial \mathcal{L}}{\partial \hat{y}_{ki}} \cdot \frac{\partial \hat{y}_{ki}}{\partial z_{ki}},
\end{equation}
where the first term is the derivative of the loss \wrt the sigmoid/softmax normalized predictions, and the second term is the derivative of the normalized predictions \wrt the unnormalized predictions (the logits $\mathbf{Z}$).
Softmax normalization couples the gradients at a particular pixel location across all classes

\begin{equation}
    \frac{\partial \hat{y}_{ki}}{\partial z_{ki}} = \hat{y}_{ki} (\delta_{ck,i} - \hat{y}_{ci}),
\end{equation}
where $\delta_{ck,i}$ is the Kronecker delta, which is 1 if $k=c$ at pixel $i$ and 0 otherwise.
This is because each individual value will depend on the logits at that pixel location across all classes.
Sigmoid normalization, in contrast, does not have this coupling effect, as sigmoid is applied individually and independently to each logit:

\begin{equation}
    \frac{\partial \hat{y}_{ki}}{\partial z_{ki}} = \hat{y}_{ki} (1 - \hat{y}_{ki}).
\end{equation}
The derivative of Cross entropy loss is

\begin{equation} \label{eq:derivative_ce}
    \frac{\partial \mathcal{L}_{CE}}{\partial \hat{y}_{ki}} = -\frac{y_{ki}}{\hat{y}_{ki}},
\end{equation}
where we can observe that the gradient at a pixel location depends on the values of the prediction and the ground truth at that pixel.
The derivative of Focal loss is

\begin{equation}
\frac{\partial \mathcal{L}_{Focal}}{\partial \hat{y}_{ki}} = \alpha y_{ki} \left[ \gamma (1 - \hat{y}_{ki})^{\gamma-1} \log(\hat{y}_{ki}) - \frac{(1 - \hat{y}_{ki})^\gamma}{\hat{y}_{ki}} \right],
\end{equation}
where $\alpha$ weighs the overall contribution of Focal loss, and $\gamma$ controls the penalization over uncertain predictions.
As a loss function modified from Cross entropy, the gradients provided by Focal loss at a particular pixel also depend on the pixel and the ground truth itself, plus on the two other hyper-parameters ($\alpha, \gamma$).
The derivative of Dice loss is

\begin{equation} \label{eq:derivative_dice}
\frac{\partial \mathcal{L}_{Dice}}{\partial \hat{y}_{ki}} = -\frac{2}{CB_k^2} (y_{ki}B_k - A_k),
\end{equation}
where $C$ is the number of classes, and $B_k$ and $A_k$ are the union and intersection, respectively, between predictions and ground truth of class $k$.
In \Cref{eq:derivative_dice}, we can observe that the gradients from Dice loss (similarly to other metric-based loss functions such as Tversky loss \cite{salehi2017tversky}) do not depend on the pixel prediction but on the ground truth at the pixel itself and on global statistics involving the number of true/false positives and true/false negatives.

Note how optimizing jointly Cross entropy and Dice loss applied to softmax probability values, which is the most common approach in domains such as in medical image segmentation, couples 1) pixel predictions across all classes (\ie, $\hat{y}_{ci} \forall c \in \{1,\dots,C\}$), 2) their ground truth ($y_{ki}$), and 3) global statistics ($A, B$ in \cref{eq:derivative_dice}).
Optimizing a loss function with the SCNP logits will, in addition, couple predictions from neighboring pixels across all classes by adding the following term to the chain rule:

\begin{equation}
    \frac{\partial \tilde{z}_{ki} }{\partial z_{ki}} = \sum_{j \in \Omega(i)} \mathbf{1}[z_{ki} = \tilde{z}_{kj}] \frac{\partial \mathcal{L}}{\partial \tilde{z}_{kj}},
\end{equation}
where $\mathbf{1}[\cdot]$ is 1 if the inner condition is true and 0 otherwise.

\subsection{Example}
Let us compute the gradients at the pixel corresponding to the yellow logits (\Cref{fig:contribution}, Paper).
With $\frac{\partial \mathcal{L}_{\overline{CE}}}{\partial \tilde{z}_{ki}} = \hat{y}_{ki} - y_{ki}$, the gradients at $ki$ are:

\begin{equation}
\begin{aligned} 
\nabla_{z_{ki}} \mathcal{L}_{CE} = [ & 
(.50 - 1) + (.56 - 1), \\ 
& (.23 - 0) + (.21 - 0) + (.19 - 0), \\ 
& (.20 - 0) + (.24 - 0) ] \\
= [ & -.94, .63, .44 ]
\end{aligned} 
\end{equation}

\section{SCNP's negligible delay in convergence} \label{app:convergence}

\begin{figure}[hbt!]
  \centering
  \includegraphics[width=0.48\textwidth]{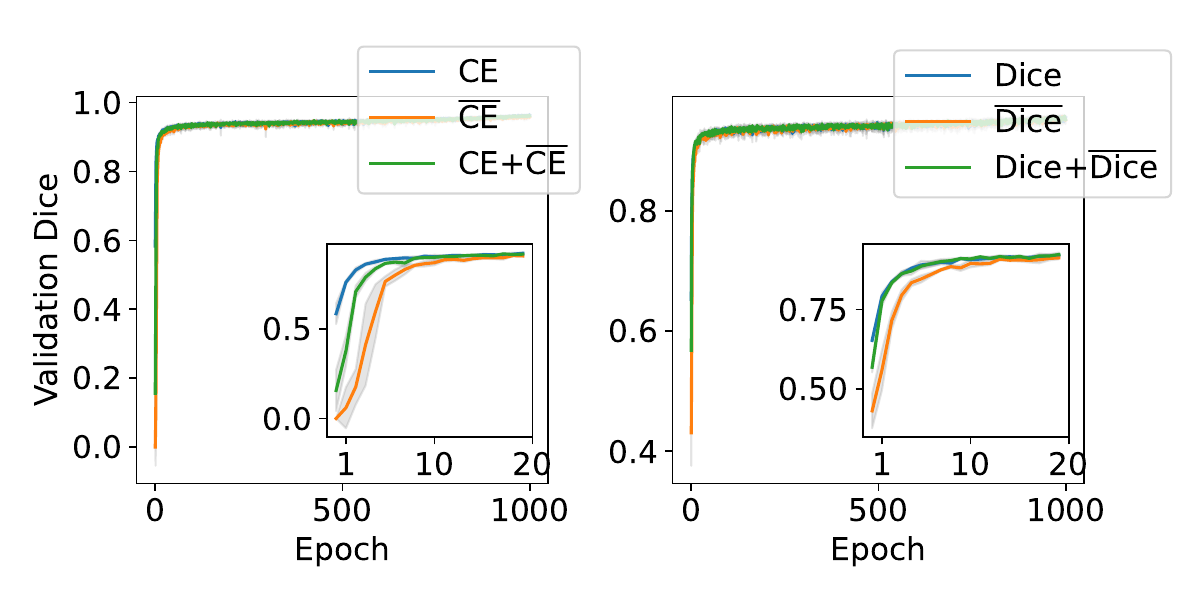}
  \caption{Left: Convergence plot when optimizing Cross entropy over the standard logits, the SCNP logits, and both. Right: Same scenario with Dice loss. The small plots inside the two panels are a zoomed in on the first 20 epochs.} \label{fig:sup_validation}
\end{figure}

Since SCNP switches the focus of the optimization from improving all pixels to improving the worst neighbors, we investigated whether and to what extent SCNP delays convergence.
\Cref{fig:sup_validation} shows, in two loss functions (Cross Entropy and Dice loss), 1) that the delay in convergence due to optimizing the SCNP logits alone is negligible, and 2) that such delay can be further reduced by optimizing the SCNP logits jointly with the standard logits. 

\section{SCNP vs. Binary closing} \label{app:binaryclosing}
We investigated the effectiveness of binary closing (•) by applying it to CEDice segmentations on four datasets (one per group). We used kernels of size 3, 5, 7 (2D \& 3D), and we applied binary closing until convergence. Below, we report the best $\beta_{0e}$ (\textit{i.e.} accuracy in the number of structures), which indicates topology accuracy.

\begin{table}[h]
\centering
{\footnotesize
\begin{tabular}{|l|lll|}
\hline
Dataset & CEDice & •(CEDice) & \textbf{SCNP} ($\overline{CEDice}$) \\ \hline
FIVES & 10.82\std{2.71} & 8.79\std{2.47} & \textbf{6.94\std{1.47}} \\
DeepGlobeRoads & 11.77\std{3.17} & 9.92\std{2.83} & \textbf{6.56\std{2.25}} \\
CirrMRI600 & 0.27\std{0.26} & 0.19\std{0.13} & \textbf{0.12\std{0.05}} \\
NuInsSeg & 8.18\std{2.39} & 8.17\std{2.40} & \textbf{7.53\std{2.20}} \\
\hline
\end{tabular}
} \caption{Average $\beta_{0e}$ in four datasets when optimizing CEDice (second column) and after applying binary closing (third column), and when optimizing CEDice with the SCNP logits (fourth column). •: Binary closing.} \label{table:app_morph} \label{tablerebuttal}
\end{table}

\noindent
Although binary closing improved $\beta_{0e}$, SCNP remained superior.
Furthermore, note that binary closing is a hurdle or ineffective in instance segmentation, real-time applications, tubular structures that are very close, and it requires a ground truth to find the optimal kernel size and number of iterations.

\section{Comparison with clCE}
\begin{figure*} 
  \centering
  \includegraphics[width=\textwidth]{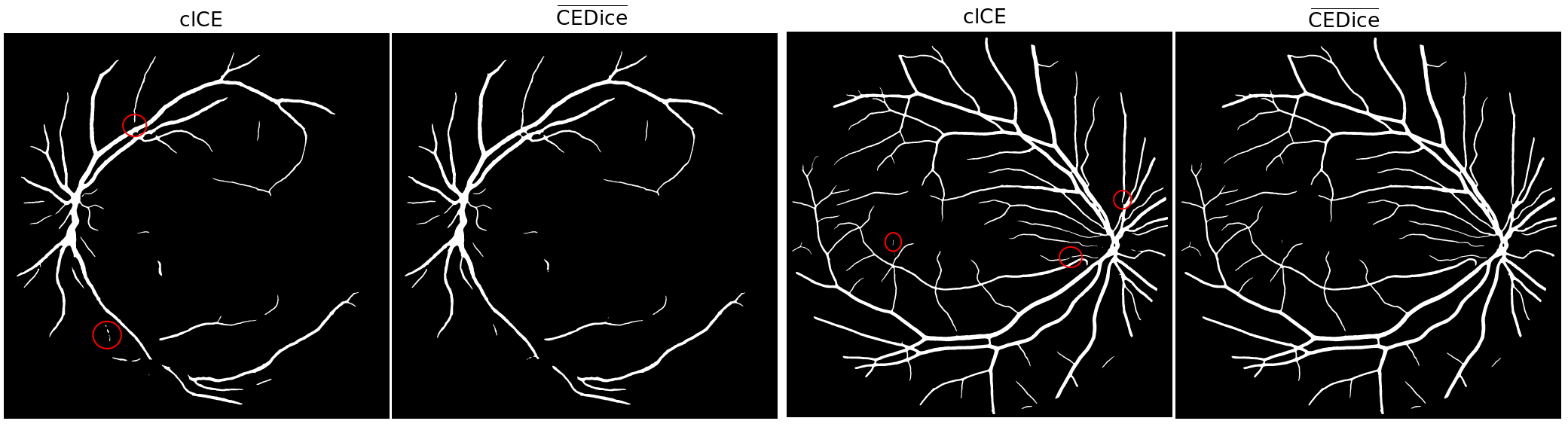}
  \caption{Visual comparison between clCE and $\overline{CEDice}$. Red circle: topological errors.} \label{figapp:clce}
\end{figure*}

We run extra experiments with \cite{acebes2024centerline}'s loss on FIVES, as FIVES was also used in \cite{acebes2024centerline}.
clCE loss achieved: Dice 0.93\std{0.00}, $\beta_{0e}$ 10.62\std{3.05}, clDice 0.92\std{0.00}, with our SCNP remaining more effective (same Dice, clDice, $\beta_{0e}$: \textbf{6.94\std{1.47}}). \Cref{figapp:clce} shows a visual comparison in two representative images.

\section{Datasets description} \label{app:datasets}
\Cref{app:datasets_tables_1,app:datasets_tables_2,app:datasets_tables_3,app:datasets_tables_4} list the details of the datasets used in this study, including an image/slice of each dataset and their size and split.


\begin{table*}[h!]
\centering
\begin{tabular}{p{0.2cm} m{5.2cm} m{4cm} m{4cm}}
\hline
\makecell[c]{\rotatebox[origin=c]{90}{FIVES}} &
\adjustbox{valign=c}{\includegraphics[width=5cm,height=5cm]{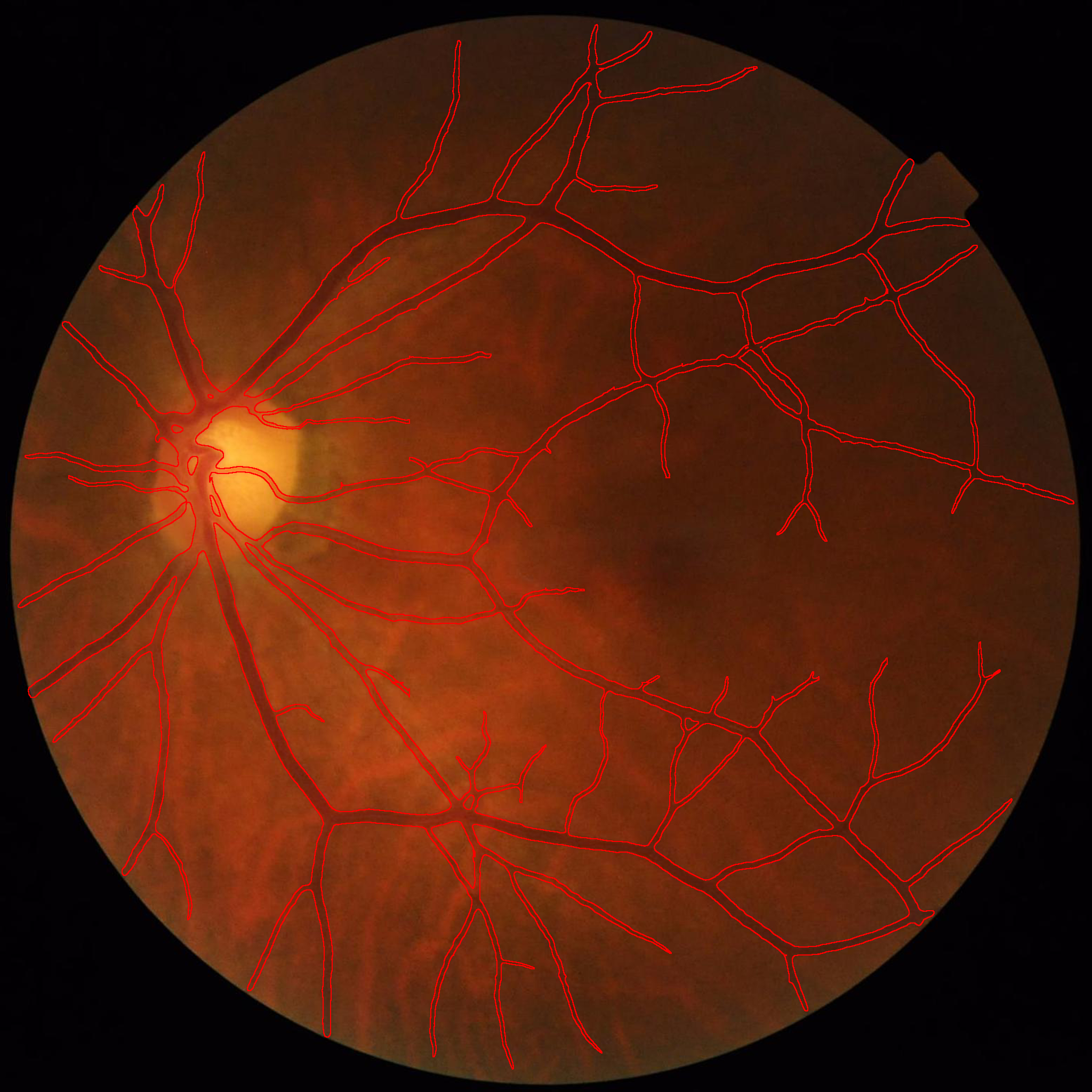}} & 
\adjustbox{valign=c}{%
\begin{minipage}[c]{\linewidth}
\begin{itemize}[leftmargin=*,nosep]
\item \underline{Blood vessels} in fundus retina images
\item Semantic segmentation
\item R,G,B
\item 2048 $\times$ 2048 pixels
\end{itemize}
\end{minipage}} &
\adjustbox{valign=c}{%
\begin{minipage}[c][4cm][t]{\linewidth}
\vspace{0.6cm}
Splits: 480-120-200

(train, validation, test)
\end{minipage}} \\
\hline
\makecell[c]{\rotatebox[origin=c]{90}{PulmonaryVA}} &
\adjustbox{valign=c}{\includegraphics[width=5cm,height=5cm]{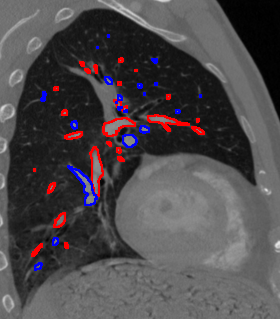}} & 
\adjustbox{valign=c}{%
\begin{minipage}[c]{\linewidth}
\begin{itemize}[leftmargin=*,nosep]
\item \underline{Pulmonary arteries and} \underline{veins} in chest CT scans
\item Semantic segmentation
\item CT
\item 184 $\times$ 235 $\times$ 262.5 voxels (median)
\end{itemize}
\end{minipage}} &
\adjustbox{valign=c}{%
\begin{minipage}[c][4cm][t]{\linewidth}
\vspace{0.6cm}
Dataset size: 106 

(five-fold cross validation)
\end{minipage}} \\
\hline
\makecell[c]{\rotatebox[origin=c]{90}{Axons}} &
\adjustbox{valign=c}{\includegraphics[width=5cm,height=5cm]{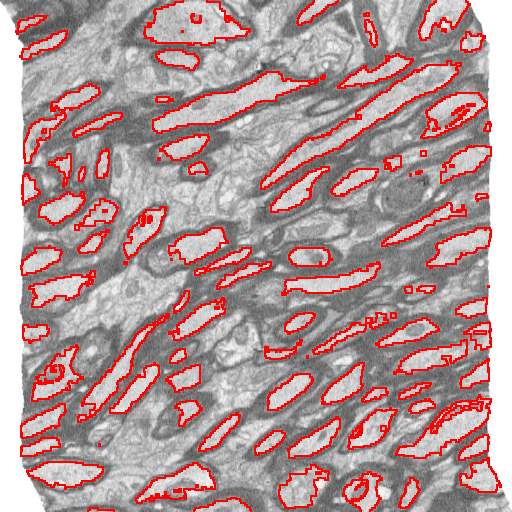}} & 
\adjustbox{valign=c}{%
\begin{minipage}[c]{\linewidth}
\begin{itemize}[leftmargin=*,nosep]
\item \underline{Axons} in mouse brain
\item Semantic segmentation
\item Electron-microscopy
\item 1062.5 $\times$ 1067 $\times$ 285 voxels (median)
\end{itemize}
\end{minipage}} &
\adjustbox{valign=c}{%
\begin{minipage}[c][4cm][t]{\linewidth}
\vspace{0.6cm}
Dataset size: 10

(five-fold cross validation)

Sham\_HM\_25\_contra|ipsi, Sham\_HM\_49\_contra|ipsi, TBI\_HM\_24\_contra|ipsi, TBI\_HM\_28\_contra|ipsi, TBI\_HM\_2\_contra|ipsi
\end{minipage}} \\
\hline
\end{tabular}
\caption{Medical datasets with tubular structures (nnUNetv2 framework, Group \ding{172})} \label{app:datasets_tables_1}
\end{table*}

\newpage

\begin{table*}[h!]
\centering
\begin{tabular}{p{0.2cm} m{5.2cm} m{4cm} m{4cm}}
\hline
\makecell[c]{\rotatebox[origin=c]{90}{TopoMortar}} &
\adjustbox{valign=c}{\includegraphics[width=5cm,height=5cm]{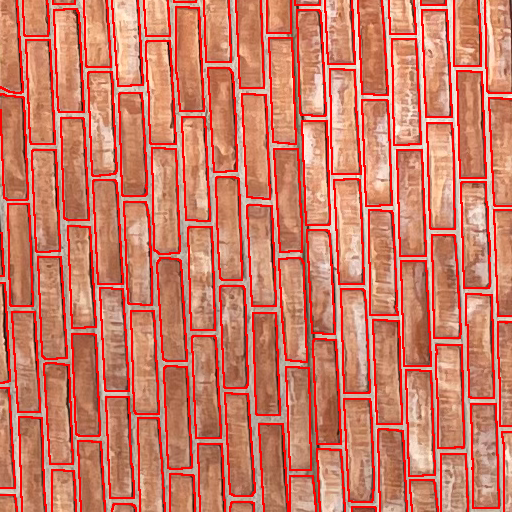}} & 
\adjustbox{valign=c}{%
\begin{minipage}[c]{\linewidth}
\begin{itemize}[leftmargin=*,nosep]
\item \underline{Mortar} in red brick wall images
\item Semantic segmentation
\item R,G,B
\item 512 $\times$ 512 pixels
\end{itemize}
\end{minipage}} &
\adjustbox{valign=c}{%
\begin{minipage}[c][4cm][t]{\linewidth}
\vspace{0.6cm}
Splits: 50-20-350

(train, validation, test)
\end{minipage}} \\
\hline
\makecell[c]{\rotatebox[origin=c]{90}{Crack500}} &
\adjustbox{valign=c}{\includegraphics[width=5cm,height=5cm]{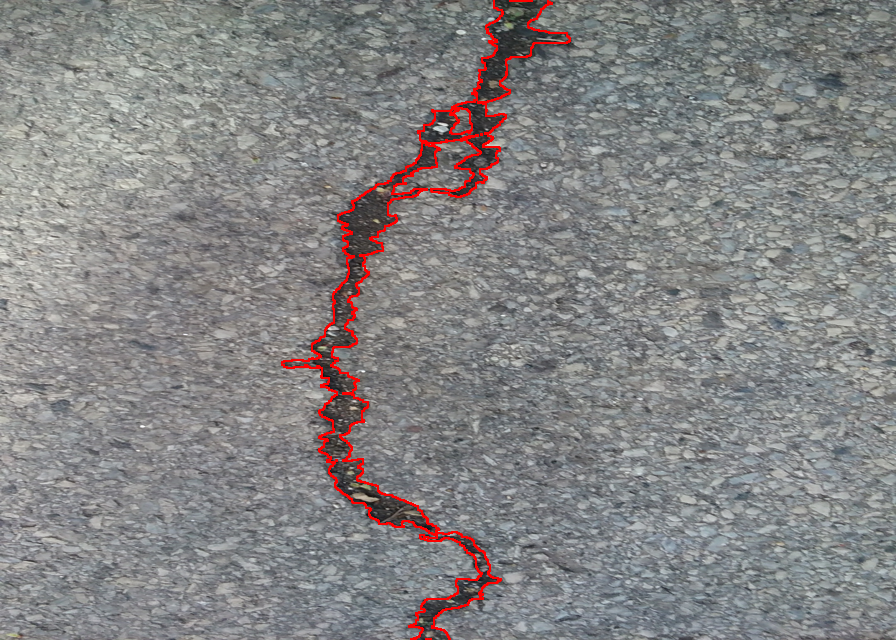}} & 
\adjustbox{valign=c}{%
\begin{minipage}[c]{\linewidth}
\begin{itemize}[leftmargin=*,nosep]
\item \underline{Cracks} in concrete images
\item Semantic segmentation
\item R,G,B
\item 640 $\times$ 896 Pixels (median)
\end{itemize}
\end{minipage}} &
\adjustbox{valign=c}{%
\begin{minipage}[c][4cm][t]{\linewidth}
\vspace{0.6cm}
Splits: 250-50-200

(train, validation, test)
\end{minipage}} \\
\hline
\makecell[c]{\rotatebox[origin=c]{90}{DeepGlobeRoads}} &
\adjustbox{valign=c}{\includegraphics[width=5cm,height=5cm]{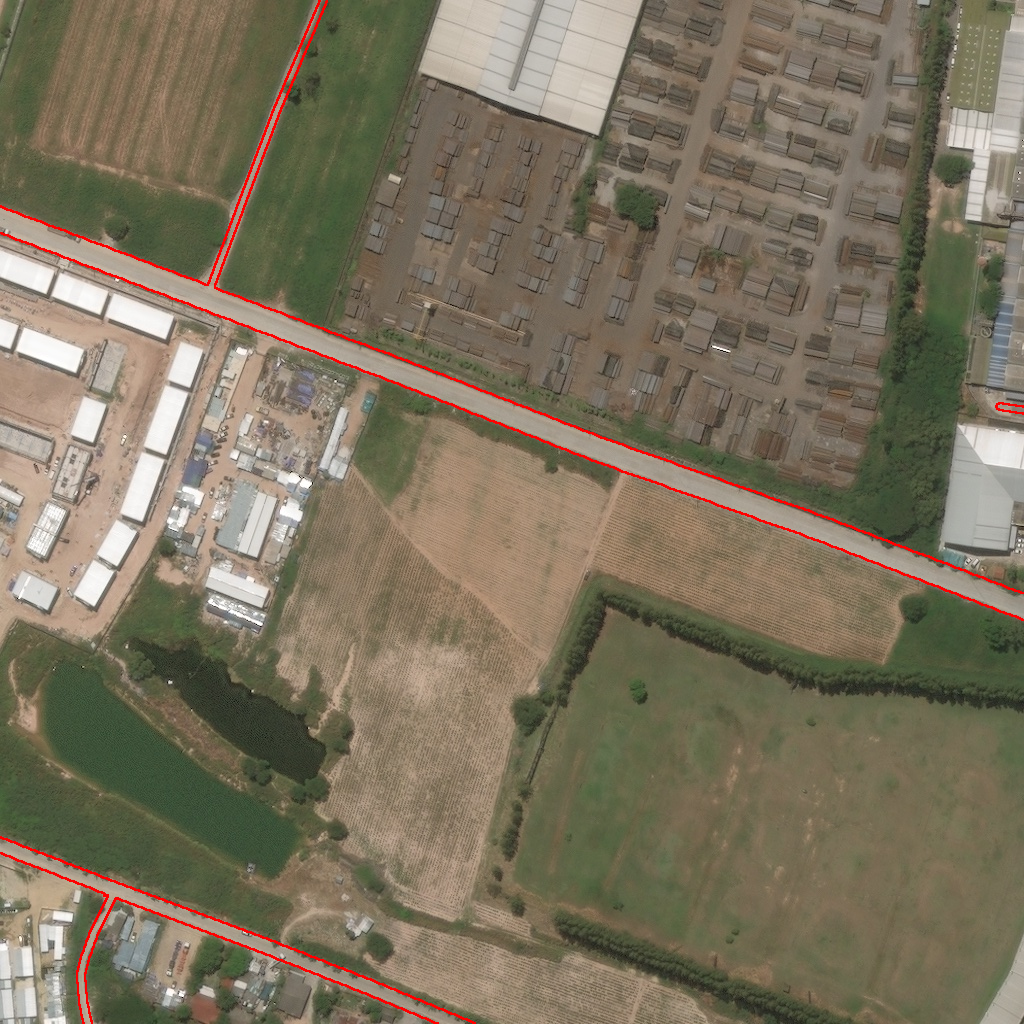}} & 
\adjustbox{valign=c}{%
\begin{minipage}[c]{\linewidth}
\begin{itemize}[leftmargin=*,nosep]
\item \underline{Roads} in satellite images
\item Semantic segmentation
\item R,G,B
\item 1024 $\times$ 1024 pixels (median)
\end{itemize}
\end{minipage}} &
\adjustbox{valign=c}{%
\begin{minipage}[c][4cm][t]{\linewidth}
\vspace{0.6cm}
Splits: 4357-623-1246

(train, validation, test)
\end{minipage}} \\
\hline
\end{tabular}
\caption{Non-medical datasets with tubular structures (Detectron2+DeepLabv3+ framework, Group \ding{173})}
\label{app:datasets_tables_2}
\end{table*}

\newpage

\begin{table*}[h!]
\centering
\begin{tabular}{p{0.2cm} m{5.2cm} m{4cm} m{4cm}}
\hline
\makecell[c]{\rotatebox[origin=c]{90}{CirrMRI600}} &
\adjustbox{valign=c}{\includegraphics[width=5cm,height=5cm]{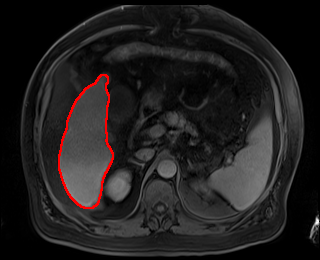}} & 
\adjustbox{valign=c}{%
\begin{minipage}[c]{\linewidth}
\begin{itemize}[leftmargin=*,nosep]
\item \underline{Cirrhotic liver} in MRI
\item Semantic segmentation
\item T2w
\item 320 $\times$ 260 $\times$ 80 voxels (median)
\end{itemize}
\end{minipage}} &
\adjustbox{valign=c}{%
\begin{minipage}[c][4cm][t]{\linewidth}
\vspace{0.6cm}
Splits: 198-50-31

(train, validation, test)
\end{minipage}} \\
\hline
\makecell[c]{\rotatebox[origin=c]{90}{ATLAS2}} &
\adjustbox{valign=c}{\includegraphics[width=5cm,height=5cm]{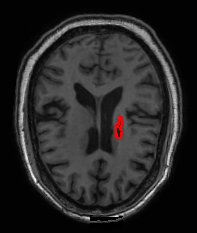}} & 
\adjustbox{valign=c}{%
\begin{minipage}[c]{\linewidth}
\begin{itemize}[leftmargin=*,nosep]
\item \underline{Lesion} in brain MRI
\item Semantic segmentation
\item T1w
\item 197 $\times$ 233 $\times$ 189 voxels (median)
\end{itemize}
\end{minipage}} &
\adjustbox{valign=c}{%
\begin{minipage}[c][4cm][t]{\linewidth}
\vspace{0.6cm}
Dataset size: 655

(five-fold cross validation)
\end{minipage}} \\
\hline
\makecell[c]{\rotatebox[origin=c]{90}{ISLES24}} &
\adjustbox{valign=c}{\includegraphics[width=5cm,height=5cm]{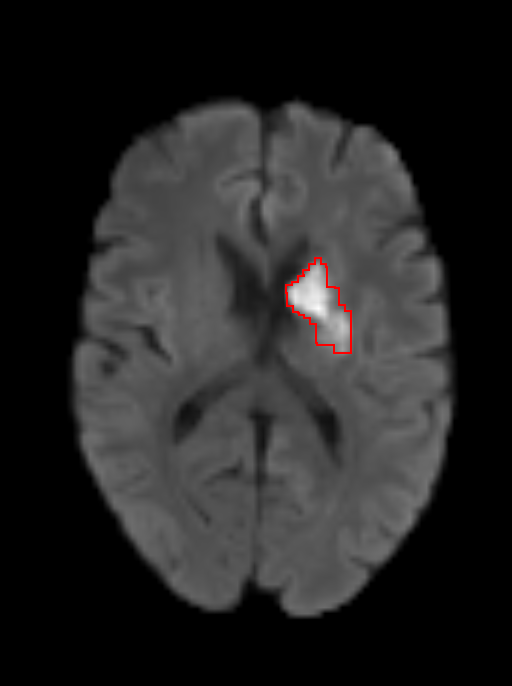}} & 
\adjustbox{valign=c}{%
\begin{minipage}[c]{\linewidth}
\begin{itemize}[leftmargin=*,nosep]
\item \underline{Ischemic lesion} in brain MRI
\item Semantic segmentation
\item ADC, DWI
\item 512 $\times$ 581 $\times$ 69 voxels (median)
\end{itemize}
\end{minipage}} &
\adjustbox{valign=c}{%
\begin{minipage}[c][4cm][t]{\linewidth}
\vspace{0.6cm}
Dataset size: 149

(five-fold cross validation)
\end{minipage}} \\
\hline
\makecell[c]{\rotatebox[origin=c]{90}{MSLesSeg}} &
\adjustbox{valign=c}{\includegraphics[width=5cm,height=5cm]{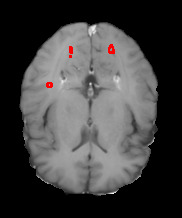}} & 
\adjustbox{valign=c}{%
\begin{minipage}[c]{\linewidth}
\begin{itemize}[leftmargin=*,nosep]
\item \underline{Multiple sclerosis lesion} in brain MRI
\item Semantic segmentation
\item T1w, T2w, FLAIR
\item 182 $\times$ 218 $\times$ 182 voxels (median)
\end{itemize}
\end{minipage}} &
\adjustbox{valign=c}{%
\begin{minipage}[c][4cm][t]{\linewidth}
\vspace{0.6cm}
Splits: 74-19-22

(train, validation, test)
\end{minipage}} \\
\hline
\end{tabular}
\caption{Medical datasets with non-tubular structures (nnUNetv2 framework, Group \ding{174})}
\label{app:datasets_tables_3}
\end{table*}


\begin{table*}[h!]
\centering
\begin{tabular}{p{0.2cm} m{5.2cm} m{4cm} m{4cm}}
\hline
\makecell[c]{\rotatebox[origin=c]{90}{NuInsSeg}} &
\adjustbox{valign=c}{\includegraphics[width=5cm,height=5cm]{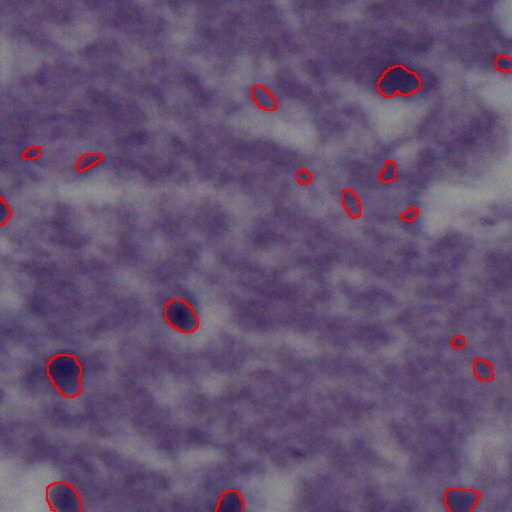}} & 
\adjustbox{valign=c}{%
\begin{minipage}[c]{\linewidth}
\begin{itemize}[leftmargin=*,nosep]
\item \underline{Cell nuclei} in H\&E-stained histological images
\item Instance segmentation
\item R,G,B
\item 512 $\times$ 512 pixels (median)
\end{itemize}
\end{minipage}} &
\adjustbox{valign=c}{%
\begin{minipage}[c][4cm][t]{\linewidth}
\vspace{0.6cm}
Splits: 532-66-67

(train, validation, test)
\end{minipage}} \\
\hline
\makecell[c]{\rotatebox[origin=c]{90}{IHC\_TMA}} &
\adjustbox{valign=c}{\includegraphics[width=5cm,height=5cm]{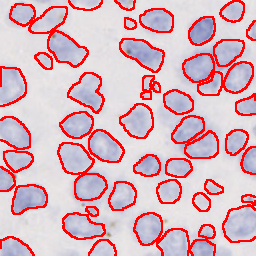}} & 
\adjustbox{valign=c}{%
\begin{minipage}[c]{\linewidth}
\begin{itemize}[leftmargin=*,nosep]
\item \underline{Cell nuclei} in immunohistochemical tissue microarray images
\item Instance segmentation
\item R,G,B
\item 256 $\times$ 256 pixels (median)
\end{itemize}
\end{minipage}} &
\adjustbox{valign=c}{%
\begin{minipage}[c][4cm][t]{\linewidth}
\vspace{0.6cm}
Splits: 212-27-27 

(train, validation, test)
\end{minipage}} \\
\hline
\makecell[c]{\rotatebox[origin=c]{90}{LyNSeC}} &
\adjustbox{valign=c}{\includegraphics[width=5cm,height=5cm]{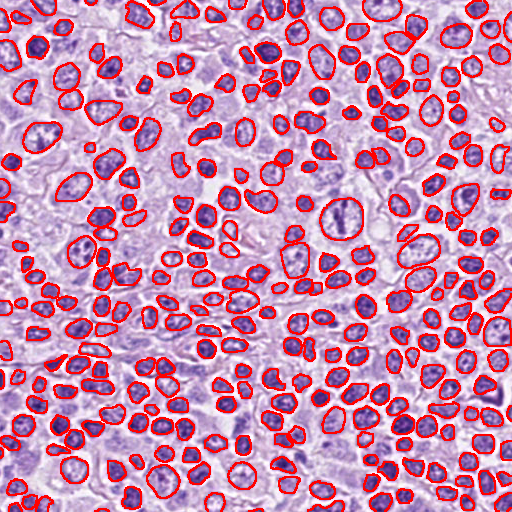}} & 
\adjustbox{valign=c}{%
\begin{minipage}[c]{\linewidth}
\begin{itemize}[leftmargin=*,nosep]
\item \underline{Lymphoma nuclei} in H\&E-stained histological images
\item Instance segmentation
\item R,G,B
\item 512 $\times$ 512 pixels (median)
\end{itemize}
\end{minipage}} &
\adjustbox{valign=c}{%
\begin{minipage}[c][4cm][t]{\linewidth}
\vspace{0.6cm}
Splits: 559-70-70

(train, validation, test)
\end{minipage}} \\
\hline
\end{tabular}
\caption{Medical (cells) datasets with rounded structures (InstanSeg framework, Group \ding{175})}
\label{app:datasets_tables_4}
\end{table*}


\clearpage
\newpage

\clearpage
\newpage

\section{Optimization} \label{app:loss_functions_extra}
We left the frameworks' optimization strategy as it was by default.

\textbf{nnUNetv2} employs SGD with Nesterov momentum (0.99), a weight decay of 3e-5, initial learning rate of 0.01, and the following learning rate scheduler: $\eta_{it} = \eta_0 (\frac{1-it}{iterations})^{0.9}$. Models were optimized for 1000 epochs.

\textbf{Detectron2} employs AdamW \cite{loshchilov2017decoupled} with an initial learning rate of 0.001, $\beta_1 = 0.9, \beta_2=0.999$, a weight decay of 0.1, and cosine annealing as learning rate scheduler. Models were optimized for 12000 iterations.

\textbf{InstanSeg} employs Adam \cite{kingma2014adam} with an initial learning rate of 0.001. Models were optimized for 500, 150 and 500 epochs in NuInsSeg, IHC\_TMA, and LyNSeC, respectively.

\subsection{Loss functions and their hyper-parameters} 
\paragraph{Cross Entropy Dice loss (CEDice)} We summed Cross Entropy and Dice loss: $\mathcal{L}_{CEDice} = \mathcal{L}_{CE} + \mathcal{L}_{Dice}$.

\paragraph{clDice loss} We employed the official code provided by \citet{shit2021cldice} and, as in the original study, we combined clDice with Dice loss. Since clDice's effectiveness depends on the suitability of the hyper-parameter ``number of iterations'' ($i$) of its soft-skeletonization method, we conducted hyper-parameter optimization. For each dataset, we explored five different values that were selected based on the thickness of the tubular structures while limiting the GPU memory usage to 80GB. The values that we explored in each dataset (and the \textbf{selected} hyper-parameter) were:

\begin{itemize}
    \item FIVES: 8, 16, \textbf{24}, 32, 40.
    \item PulmonaryVA: 4, 8, 12, \textbf{16}, 18.
    \item Axons: 9, 18, 27, 36, \textbf{46}.
    \item TopoMortar: \textbf{4}, 8, 12, 16, 18.
    \item Crack500: \textbf{10}, 20, 30, 40, 51.
    \item DeepGlobeRoads: 15, 30, \textbf{45}, 60, 72.
\end{itemize}

\paragraph{Skeleton Recall (SkelRecall)} We employed the official code provided by \citet{kirchhoff2024skeleton} and, as in the original study, we combined SkelRecall with Cross Entropy loss.

\paragraph{TopoLoss} We employed the official code provided by \citet{hu2019topology}.
As suggested in the original work, we first optimized Cross Entropy loss and, at the end of the optimization, we combined it with TopoLoss.
Since TopoLoss computes persistence homology, the iteration time increased dramatically, especially for the 3D datasets.
Therefore, we set the number of iterations that we optimized TopoLoss such that the total training time did not exceed 24 hours. \Cref{table:topoloss_details} summarizes this.

\begin{table}[h]
\centering
{
\begin{tabular}{|l|ll|}
\hline
Dataset & Eps. w/ TopoLoss & \% of train time \\ \hline
FIVES & 38 (1000) & 44.20\% \\
PulmonaryVA & 3 (1000) & 34.30\%  \\
Axons & 5 (1000) & 51.88\% \\
\hline
TopoMortar & 3600 (12000) & 87.86\%  \\
Crack500 & 3600 (12000) & 62.65\%  \\
DeepGlobeRoads & 3600 (12000) & 41.50\%  \\
\hline
CirrMRI600 & 5 (1000) & 39.11\%  \\
ATLAS2 & 12 (1000) & 37.16\% \\
ISLES24 & 12 (1000) & 63.35\% \\
MSLesSeg & 5 (1000) & 36.81\% \\
\hline
NuInsSeg & 150 (500) & 36.80\% \\
IHC\_TMA & 15 (150) &  34.41\% \\
LyNSeC & 150 (500) & 53.72\% \\
\hline
\end{tabular}
} \caption{Second column: Epochs optimizing TopoLoss and total epochs. Third column: Percentage of training time optimizing TopoLoss relative to the total training time (average across the five runs).} \label{table:topoloss_details}
\end{table}

\paragraph{RegionWise (RWLoss)} We employed the official code provided by \citet{valverde2023region}.

\paragraph{Focal loss} We employed MONAI's implementation \cite{cardoso2022monai} of Focal loss \cite{lin2017focal}, using its default hyper-parameters.

\paragraph{Tversky loss} We employed MONAI's implementation \cite{cardoso2022monai} of Tversky loss \cite{salehi2017tversky}, and we optimized its hyper-parameters $\alpha,\beta$ which control the penalty for false positives and negatives.

\section{Performance metrics} \label{app:main_results_tables}

\Cref{table:skeleton_medical_nnunetv2,table:skeleton_natural_detectron,table:round_medical_nnunet,table:round_cells_instance} show the results obtained by CEDice (our baseline), CEDice optimizing the SCNP-penalized logits, their combination, and several loss function, on the 13 datasets used in this study.
Furthermore, we assess the statistical significance of SCNP-derived performance metrics compared to the baseline CEDice with a permutation t-test based on 10,000 random permutations (p $<$ 0.05).

For the datasets with a pre-defined train-test split (\Cref{app:datasets}), we report the mean and standard deviation on the same test-set images across five runs; for the remaining datasets, we report the mean and standard deviation across the five-fold cross validation runs.
In consequence, the standard deviation in the first group of datasets indicates algorithmic variance while in the second group it indicates intra-dataset variance. \Cref{app:quality} presents a qualitative comparison of segmentations.

\begin{table}[h]
{\footnotesize
\begin{tabular}{|ll|lll|}
\hline
& Loss & Dice & $\beta_{0e}$ & clDice \\ \hline
\parbox[t]{2mm}{\multirow{9}{*}{\rotatebox[origin=c]{90}{\shortstack[c]{FIVES}}}} & CEDice & 0.92\std{0.0} & 10.82\std{2.71} & 0.92\std{0.0} \\ 
& clDice ($i = 24$) & 0.91\std{0.0} & 36.55\std{10.70} & 0.91\std{0.01} \\ 
& SkelRecall & 0.91\std{0.0} & 12.45\std{4.25} & 0.91\std{0.0} \\ 
& TopoLoss & 0.92\std{0.0} & 8.09\std{2.45} & 0.92\std{0.0} \\ 
& RWLoss & 0.92\std{0.0} & 15.99\std{5.35} & 0.92\std{0.0} \\ 
& Tversky ($\beta = 0.7$) & 0.92\std{0.0} & 12.22\std{3.45} & 0.92\std{0.0} \\ 
& Focal & 0.92\std{0.0} & 16.08\std{5.97} & 0.92\std{0.0} \\ 
\cline{3-5}
& CEDice+$\overline{CEDice}$ & \textbf{0.93\std{0.0}} & 7.95\std{1.82} & 0.92\std{0.0} \\ 
& $\overline{CEDice}$ & 0.92\std{0.0} & \textbf{6.94\std{1.47}} & 0.92\std{0.0} \\ 
\hline
\parbox[t]{2mm}{\multirow{9}{*}{\rotatebox[origin=c]{90}{\shortstack[c]{PulmonaryVA}}}} & CEDice & 0.84\std{0.04} & 12.89\std{12.92} & 0.88\std{0.04} \\ 
& clDice ($i = 16$) & 0.84\std{0.04} & 15.98\std{8.73} & 0.89\std{0.04} \\ 
& SkelRecall & 0.83\std{0.05} & 39.45\std{20.42} & 0.84\std{0.06} \\ 
& TopoLoss & 0.84\std{0.05} & 14.44\std{10.67} & 0.88\std{0.04} \\ 
& RWLoss & 0.84\std{0.05} & 18.02\std{13.83} & 0.87\std{0.04} \\ 
& Tversky ($\beta = 0.6$) & 0.84\std{0.05} & 15.44\std{14.87} & 0.88\std{0.04} \\ 
& Focal & 0.84\std{0.05} & 21.29\std{13.4} & 0.87\std{0.04} \\ 
\cline{3-5}
& CEDice+$\overline{CEDice}$ & 0.84\std{0.05} & \textbf{5.82\std{5.02}} & 0.87\std{0.06} \\ 
& $\overline{CEDice}$ & 0.83\std{0.05} & \textbf{6.95\std{4.18}} & 0.84\std{0.07} \\ 
\hline
\parbox[t]{2mm}{\multirow{9}{*}{\rotatebox[origin=c]{90}{\shortstack[c]{Axons}}}} & CEDice & 0.90\std{0.04} & 531\std{534} & 0.85\std{0.05} \\ 
& clDice ($i = 46$) & 0.89\std{0.04} & 1925\std{1701} & 0.86\std{0.04} \\ 
& SkelRecall & 0.89\std{0.04} & 1578\std{900} & 0.85\std{0.05} \\ 
& TopoLoss & 0.90\std{0.04} & 539\std{490} & 0.85\std{0.05} \\ 
& RWLoss & 0.90\std{0.04} & 408\std{412} & 0.85\std{0.05} \\ 
& Tversky ($\beta = 0.6$) & 0.90\std{0.04} & 530\std{623} & 0.85\std{0.05} \\ 
& Focal & 0.90\std{0.04} & 412\std{400} & 0.85\std{0.05} \\ 
\cline{3-5}
& CEDice+$\overline{CEDice}$ & 0.90\std{0.04} & 254\std{203} & 0.87\std{0.05} \\ 
& $\overline{CEDice}$ & 0.90\std{0.04} & 279\std{207} & 0.88\std{0.05} \\ 
\hline
\end{tabular}
} \caption{Group \ding{172} Medical datasets with tubular structures (nnUNetv2). $\overline{CEDice}$: CEDice applied to SCNP. Bold (Dice): Highest and significantly significant. Bold ($\beta_{0e}$, clDice): Best scores with Dice equal or better and statistically significant.} \label{table:skeleton_medical_nnunetv2}
\end{table}

\begin{table}[h]
{\footnotesize
\begin{tabular}{|ll|lll|}
\hline
& Loss & Dice & $\beta_{0e}$ & clDice \\ \hline
\parbox[t]{2mm}{\multirow{9}{*}{\rotatebox[origin=c]{90}{\shortstack[c]{TopoMortar}}}}  & CEDice & 0.67\std{0.03} &  25.06\std{9.26} & 0.75\std{0.03} \\ 
 & clDice ($i=4$) & 0.68\std{0.02} &  21.40\std{5.86} & 0.75\std{0.02} \\ 
 & SkelRecall & 0.68\std{0.02} &  23.64\std{7.30} & 0.76\std{0.02} \\ 
 & TopoLoss & 0.67\std{0.02} &  27.22\std{7.64} & 0.74\std{0.02} \\ 
 & RWLoss & 0.54\std{0.30} &  19.40\std{13.80} & 0.76\std{0.03} \\ 
 & Tversky ($\beta = 0.8$) & 0.68\std{0.02} &  22.59\std{6.46} & 0.75\std{0.02} \\ 
 & Focal & 0.67\std{0.02} &  29.75\std{10.17} & 0.74\std{0.02} \\ 
\cline{3-5}
 & CEDice+$\overline{CEDice}$ & 0.68\std{0.02} &  20.86\std{6.78} & 0.76\std{0.02} \\ 
 & $\overline{CEDice}$ & 0.68\std{0.02} &  20.30\std{6.45} & 0.75\std{0.02} \\ 
\hline
\parbox[t]{2mm}{\multirow{9}{*}{\rotatebox[origin=c]{90}{\shortstack[c]{Crack500}}}}  & CEDice & 0.68\std{0.03} &  10.61\std{3.44} & 0.76\std{0.03} \\ 
 & clDice ($i=10$) & 0.65\std{0.09} &  9.98\std{2.96} & 0.80\std{0.04} \\ 
 & SkelRecall & 0.65\std{0.02} &  11.78\std{3.22} & 0.77\std{0.02} \\ 
 & TopoLoss & 0.54\std{0.04} &  13.34\std{3.24} & 0.58\std{0.04} \\ 
 & RWLoss & --- &  --- & --- \\ 
 & Tversky ($\beta = 0.6$) & 0.69\std{0.02} &  10.47\std{3.67} & 0.77\std{0.03} \\ 
 & Focal & --- &  --- & --- \\ 
\cline{3-5}
 & CEDice+$\overline{CEDice}$ & 0.68\std{0.03} &  9.32\std{2.71} & 0.76\std{0.03} \\ 
 & $\overline{CEDice}$ & 0.66\std{0.03} &  9.48\std{2.36} & 0.73\std{0.03} \\ 
\hline
\parbox[t]{2mm}{\multirow{9}{*}{\rotatebox[origin=c]{90}{\shortstack[c]{DeepGlobeRoads}}}}  & CEDice & 0.75\std{0.02} &  11.77\std{3.17} & 0.82\std{0.02} \\ 
 & clDice ($i=45$) & 0.72\std{0.03} &  6.51\std{2.41} & 0.82\std{0.03} \\ 
 & SkelRecall & 0.71\std{0.02} &  15.26\std{4.12} & 0.79\std{0.02} \\ 
 & TopoLoss & 0.22\std{0.05} &  27.79\std{6.04} & 0.33\std{0.06} \\ 
 & RWLoss & --- & --- & --- \\ 
 & Tversky ($\beta = 0.7$) & 0.74\std{0.02} &  9.78\std{2.90} & 0.82\std{0.02} \\ 
 & Focal &--- &  --- & --- \\ 
\cline{3-5}
 & CEDice+$\overline{CEDice}$ & 0.72\std{0.03} &  7.24\std{2.63} & 0.80\std{0.03} \\ 
 & $\overline{CEDice}$ & 0.70\std{0.03} &  6.56\std{2.25} & 0.78\std{0.03} \\ 
\hline
\end{tabular}
} \caption{Group \ding{173} Non-medical datasets with tubular structures (Detectron2+DeepLabv3+).} \label{table:skeleton_natural_detectron}
\end{table}

\begin{table}[h]
\centering
{\footnotesize
\begin{tabular}{|ll|ll|}
\hline
& Loss & Dice & $\beta_{0e}$  \\ \hline
\parbox[t]{2mm}{\multirow{7}{*}{\rotatebox[origin=c]{90}{\shortstack[c]{CirrMRI600}}}} & CEDice & 0.97\std{0.01} & 0.27\std{0.26}  \\ 
& TopoLoss & 0.97\std{0.01} & 0.32\std{0.28}  \\ 
& RWLoss & 0.96\std{0.01} & 0.43\std{0.31}  \\ 
& Tversky ($\beta = 0.7$) & 0.95\std{0.0} & 1.39\std{1.49}  \\ 
& Focal & 0.97\std{0.01} & 0.51\std{0.30}  \\ 
\cline{3-4}
& CEDice+$\overline{CEDice}$ & 0.97\std{0.01} & 0.17\std{0.09}  \\ 
& $\overline{CEDice}$ & 0.97\std{0.01} & \textbf{0.12\std{0.05}}  \\ 
\hline
\parbox[t]{2mm}{\multirow{7}{*}{\rotatebox[origin=c]{90}{\shortstack[c]{ATLAS2}}}} & CEDice & 0.63\std{0.28} & 1.60\std{2.5}  \\ 
& TopoLoss & 0.61\std{0.30} & 1.45\std{2.42}  \\ 
& RWLoss & 0.56\std{0.33} & 1.44\std{2.47}  \\ 
& Tversky ($\beta = 0.8$) & 0.61\std{0.26} & 2.37\std{3.15}  \\ 
& Focal & 0.59\std{0.32} & 1.39\std{2.51}  \\ 
\cline{3-4}
& CEDice+$\overline{CEDice}$ & 0.62\std{0.27} & 1.40\std{2.49}  \\ 
& $\overline{CEDice}$ & 0.61\std{0.28} & 1.38\std{2.57}  \\ 
\hline
\parbox[t]{2mm}{\multirow{7}{*}{\rotatebox[origin=c]{90}{\shortstack[c]{ISLES24}}}} & CEDice & 0.77\std{0.20} & 3.64\std{4.39}  \\ 
& TopoLoss & 0.77\std{0.20} & 3.77\std{5.06}  \\ 
& RWLoss & 0.77\std{0.20} & 3.92\std{5.03}  \\ 
& Tversky ($\beta = 0.7$) & 0.77\std{0.19} & 4.61\std{7.65}  \\ 
& Focal & 0.76\std{0.21} & 4.32\std{6.30}  \\ 
\cline{3-4}
& CEDice+$\overline{CEDice}$ & 0.77\std{0.19} & 3.72\std{5.40}  \\ 
& $\overline{CEDice}$ & 0.76\std{0.20} & 5.27\std{7.85}  \\ 
\hline
\parbox[t]{2mm}{\multirow{7}{*}{\rotatebox[origin=c]{90}{\shortstack[c]{MSLesSeg}}}} & CEDice & 0.71\std{0.01} & 9.47\std{1.81}  \\ 
& TopoLoss & 0.71\std{0.01} & 9.40\std{1.84}  \\ 
& RWLoss & 0.70\std{0.01} & 10.22\std{1.64}  \\ 
& Tversky ($\beta = 0.8$) & 0.71\std{0.01} & 9.55\std{2.14}  \\ 
& Focal & 0.69\std{0.02} & 10.06\std{2.27}  \\ 
\cline{3-4}
& CEDice+$\overline{CEDice}$ & 0.70\std{0.01} & 12.72\std{1.54}  \\ 
& $\overline{CEDice}$ & 0.67\std{0.01} & 23.05\std{1.22}  \\ 
\hline
\end{tabular}
} \caption{Group \ding{174} Medical datasets with non-tubular structures (nnUNetv2).} \label{table:round_medical_nnunet}
\end{table}

\begin{table}[h]
{\footnotesize
\begin{tabular}{|ll|lll|}
\hline
& Loss & Dice & $\beta_{0e}$ & $(\mathcal{R}_e)$ \\ \hline
\parbox[t]{2mm}{\multirow{7}{*}{\rotatebox[origin=c]{90}{\shortstack[c]{NuInsSeg}}}} & CEDice & 0.76\std{0.04} & 8.18\std{2.39} &  0.91\std{0.01} \\ 
& TopoLoss & 0.28\std{0.01} & 13.35\std{9.23} &  0.76\std{0.04} \\ 
& RWLoss & 0.72\std{0.04} & 8.42\std{3.04} &  0.90\std{0.01} \\ 
& Tversky ($\beta=0.6$) & 0.77\std{0.03} & 7.94\std{2.91} &  0.90\std{0.01} \\ 
& Focal & 0.71\std{0.03} & 9.25\std{2.05} &  0.89\std{0.01} \\ 
\cline{3-5}
& CEDice+$\overline{CEDice}$ & 0.76\std{0.04} & 7.95\std{2.21} &  0.91\std{0.01} \\ 
& $\overline{CEDice}$ & 0.76\std{0.03} & 7.53\std{2.20} &  \textbf{0.92\std{0.01}} \\ 
\hline
\parbox[t]{2mm}{\multirow{7}{*}{\rotatebox[origin=c]{90}{\shortstack[c]{IHC\_TMA}}}} & CEDice & 0.79\std{0.04} & 3.31\std{1.58} &  0.88\std{0.01} \\ 
& TopoLoss & 0.36\std{0.0} & 7.49\std{3.36} &  0.78\std{0.03} \\ 
& RWLoss & 0.78\std{0.05} & 2.33\std{1.36} &  0.87\std{0.01} \\ 
& Tversky ($\beta=0.9$) & 0.81\std{0.04} & 2.64\std{1.32} &  0.88\std{0.01} \\ 
& Focal & 0.77\std{0.04} & 2.33\std{1.22} &  0.87\std{0.01} \\ 
\cline{3-5}
& CEDice+$\overline{CEDice}$ & 0.82\std{0.04} & 2.39\std{1.24} &  0.89\std{0.01} \\ 
& $\overline{CEDice}$ & 0.81\std{0.04} & 2.81\std{2.04} &  \textbf{0.90\std{0.01}} \\ 
\hline
\parbox[t]{2mm}{\multirow{7}{*}{\rotatebox[origin=c]{90}{\shortstack[c]{LyNSeC}}}} & CEDice & 0.88\std{0.0} & 10.23\std{3.94} &  0.91\std{0.0} \\ 
& TopoLoss & 0.47\std{0.0} & 34.44\std{19.91} &  0.73\std{0.05} \\ 
& RWLoss & 0.84\std{0.01} & 10.56\std{5.04} &  0.91\std{0.0} \\ 
& Tversky ($\beta=0.6$) & 0.88\std{0.0} & 9.47\std{3.53} &  0.91\std{0.0} \\ 
& Focal & 0.86\std{0.01} & 10.10\std{3.92} &  0.90\std{0.0} \\ 
\cline{3-5}
& CEDice+$\overline{CEDice}$ & 0.88\std{0.0} & 9.75\std{3.42} &  \textbf{0.92\std{0.0}} \\ 
& $\overline{CEDice}$ & 0.88\std{0.0} & 9.91\std{3.33} &  \textbf{0.92\std{0.0}} \\ 
\hline
\end{tabular}
} \caption{ Group \ding{175} Medical datasets with rounded structures.} 
\label{table:round_cells_instance}
\end{table}

\clearpage
\newpage

\section{Segmentation examples} \label{app:quality}
\begin{figure*} 
  \centering
  \includegraphics[width=\textwidth]{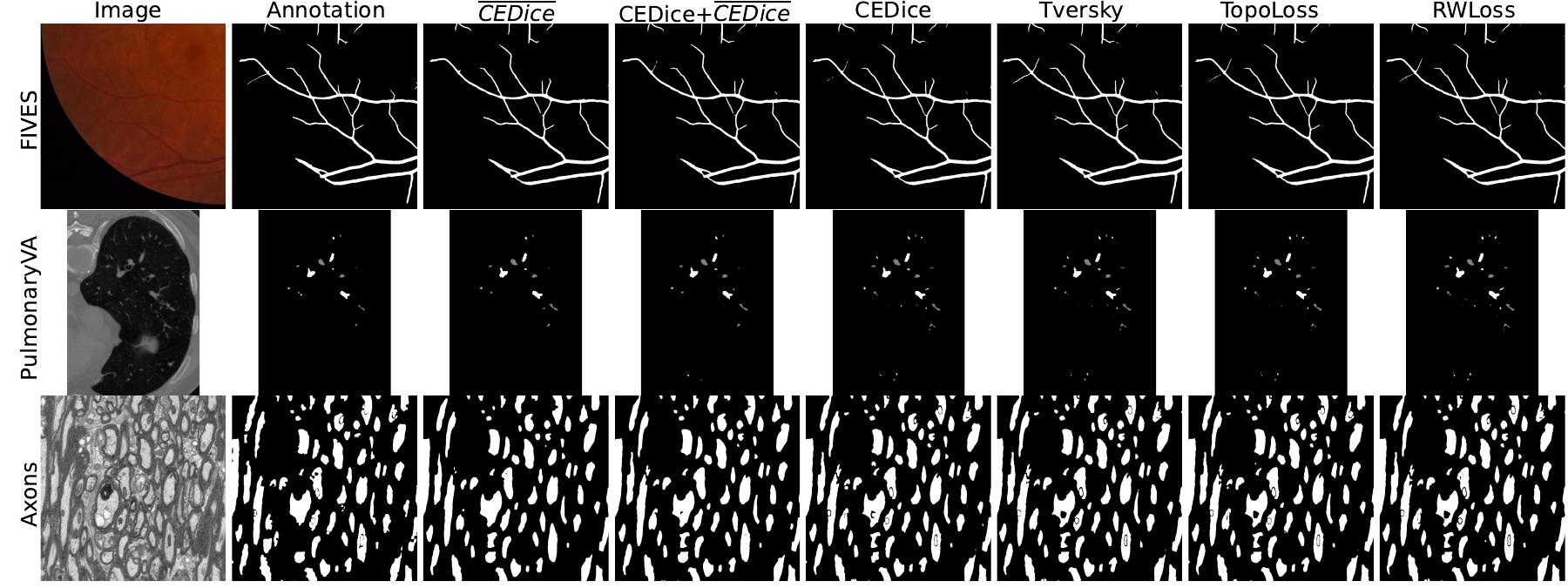}
  \caption{Results on medical datasets with tubular structures (Group \ding{172}).} \label{fig:app_results_1}
\end{figure*}

\begin{figure*} 
  \centering
  \includegraphics[width=\textwidth]{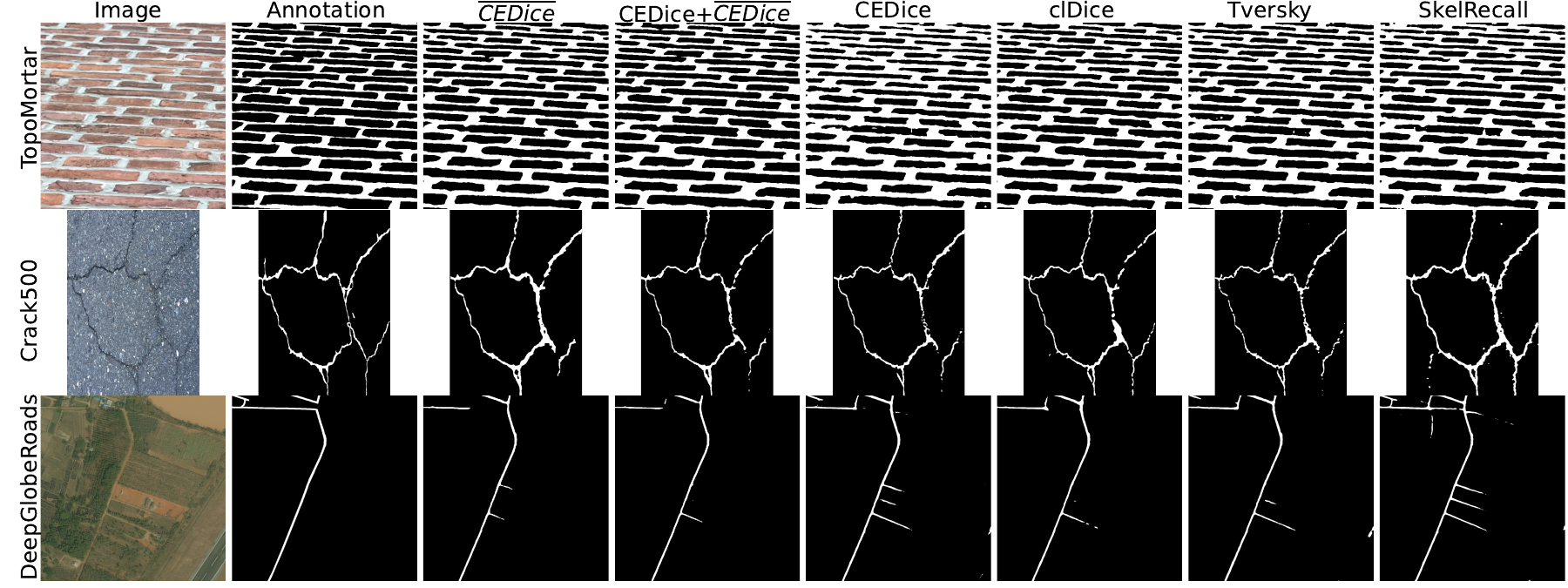}
  \caption{Results on non-medical datasets with tubular structures (Group \ding{173}).} \label{fig:app_results_2}
\end{figure*}

\begin{figure*} 
  \centering
  \includegraphics[width=\textwidth]{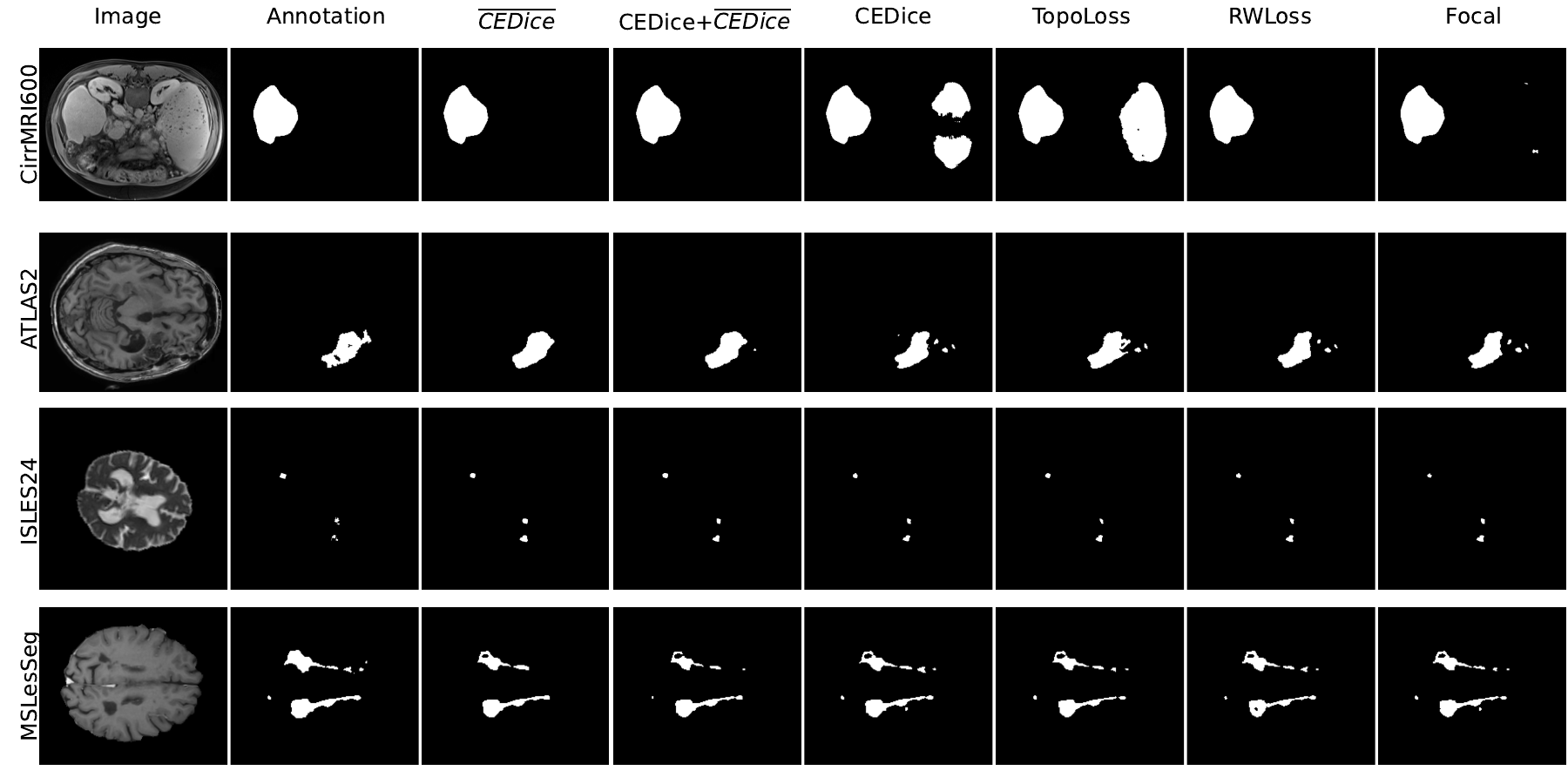}
  \caption{Results on medical datasets with non-tubular structures (Group \ding{174}).} \label{fig:app_results_3}
\end{figure*}

\begin{figure*} 
  \centering
  \includegraphics[width=\textwidth]{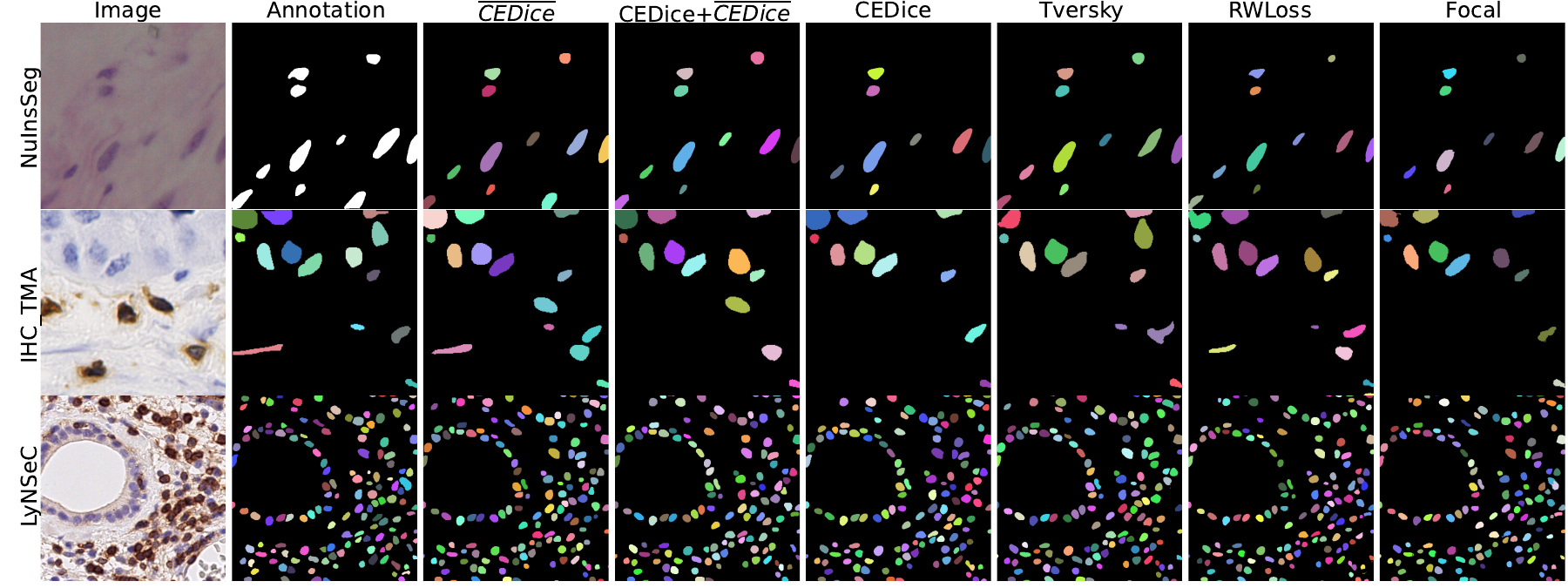}
  \caption{Results on medical datasets with rounded structures (Group \ding{175}).} \label{fig:app_results_4}
\end{figure*}

\Cref{fig:app_results_1,fig:app_results_2,fig:app_results_3,fig:app_results_4} illustrates the segmentations provided by our SCNP and other loss functions.
For each group we show the segmentations obtained by optimizing CEDice loss and the top three loss functions.


\section{Sensitivity analysis: extra metrics} \label{app:sensitivity_dice_cldice}
\Cref{table:sup_sens_dice} and \Cref{table:sup_sens_cldice} show the Dice and clDice metrics of the sensitivity analysis.

\begin{table}[hbt!]
\centering
{\small
\begin{tabular}{|l|lll|}
\hline
$\downarrow w$ & 9.69 & 7.23 & 4.70 \\ \hline
3 & 0.92\std{0.0} & 0.92\std{0.0} & 0.92\std{0.0} \\
5 & 0.92\std{0.0} & 0.92\std{0.0} & 0.92\std{0.0} \\
7 & 0.92\std{0.0} & 0.91\std{0.0} & 0.91\std{0.0} \\
9 & 0.9\std{0.0} & 0.89\std{0.0} & 0.9\std{0.0} \\
11 & 0.92\std{0.0} & 0.91\std{0.0} & 0.92\std{0.0} \\
\hline
\end{tabular}
} \caption{Dice coefficients of the sensitivity analysis on SCNP's hyper-parameter $w$.} \label{table:sup_sens_dice}
\end{table}

\begin{table}[hbt!]
\centering
{\small
\begin{tabular}{|l|lll|}
\hline
$\downarrow w$ & 9.69 & 7.23 & 4.70 \\ \hline
3 & 0.92\std{0.0} & 0.92\std{0.0} & 0.92\std{0.0} \\
5 & 0.91\std{0.0} & 0.91\std{0.0} & 0.92\std{0.0} \\
7 & 0.9\std{0.0} & 0.89\std{0.0} & 0.9\std{0.0} \\
9 & 0.88\std{0.0} & 0.87\std{0.0} & 0.88\std{0.0} \\
11 & 0.91\std{0.0} & 0.91\std{0.0} & 0.92\std{0.0} \\
\hline
\end{tabular}
} \caption{clDice metric of the sensitivity analysis on SCNP's hyper-parameter $w$.} \label{table:sup_sens_cldice}
\end{table}

\end{document}